\documentclass{article}

\usepackage[preprint]{neurips_2024}

\usepackage[utf8]{inputenc} % allow utf-8 input
\usepackage[T1]{fontenc}    % use 8-bit T1 fonts
\usepackage{hyperref}       % hyperlinks
\usepackage{url}            % simple URL typesetting
\usepackage{xurl}  % 自动换行网址
\usepackage{booktabs}       % professional-quality tables
\usepackage{amsfonts}       % blackboard math symbols
\usepackage{nicefrac}       % compact symbols for 1/2, etc.
\usepackage{microtype}      % microtypography
\usepackage{xcolor}         % colors

\usepackage{amsthm}
\usepackage{amsbsy}
\usepackage{amsmath}
\usepackage{amssymb}

\usepackage{algorithm}
\usepackage{algpseudocode} % 或者使用 algorithmic

\usepackage{graphicx}
\usepackage{natbib}
\usepackage{cleveref}
\usepackage{wrapfig}
\usepackage{subfig}

\usepackage{multirow}
\usepackage{diagbox}

\usepackage{pifont}
\usepackage{enumitem}

\title{Learnware of Language Models: \\Specialized Small Language Models Can Do Big}

\author{%
  Zhi-Hao Tan\thanks{These authors contributed equally.}, Zi-Chen Zhao\footnotemark[1], Hao-Yu Shi\footnotemark[1], Xin-Yu Zhang, Peng Tan, Yang Yu, Zhi-Hua Zhou
  \\
  National Key Laboratory for Novel Software Technology, Nanjing University, China\\
  School of Artificial Intelligence, Nanjing University, China \\
  \texttt{\{tanzh,zhaozc,shihy,zhangxinyu2024,tanp,yuy,zhouzh\}@lamda.nju.edu.cn} \\
}

\begin{document}

\maketitle

\begin{abstract}

The \emph{learnware} paradigm offers a novel approach to machine learning by enabling users to reuse a set of well-trained models for tasks beyond the models' original purposes. It eliminates the need to build models from scratch, instead relying on \emph{specifications} (representations of a model’s capabilities) to identify and leverage the most suitable models for new tasks. While learnware has proven effective in many scenarios, its application to language models has remained largely unexplored. At the same time, large language models (LLMs) have demonstrated remarkable universal question-answering abilities, yet they face challenges in specialized scenarios due to data scarcity, privacy concerns, and high computational costs, thus more and more specialized small language models (SLMs) are being trained for specific domains. 
To address these limitations systematically, the learnware paradigm provides a promising solution by enabling maximum utilization of specialized SLMs, and allowing users to identify and reuse them in a collaborative and privacy-preserving manner.
 
This paper presents a preliminary attempt to apply the learnware paradigm to language models. We simulated a learnware system comprising approximately 100 learnwares of specialized SLMs with 8B parameters, fine-tuned across finance, healthcare, and mathematics domains. Each learnware contains an SLM and a specification, which enables users to identify the most relevant models without exposing their own data. Experimental results demonstrate promising performance: by selecting one suitable learnware for each task-specific inference, the system outperforms the base SLMs on all benchmarks. Compared to LLMs, the system outperforms Qwen1.5-110B, Qwen2.5-72B, and Llama3.1-70B-Instruct by at least 14\% in finance domain tasks. Additionally, it surpasses Flan-PaLM-540B (ranked 7th on the Open Medical LLM Leaderboard) in medical domain tasks.

These preliminary results highlight the potential of the learnware paradigm to organize and utilize specialized SLMs effectively, offering a scalable, efficient, and privacy-preserving framework for solving diverse specialized tasks.

\end{abstract}

\section{Introduction}

Large language models (LLMs) have achieved remarkable performance across a wide range of general tasks. Directly applying these models in specialized scenarios still faces significant challenges. On one hand, suffering from limited or hard-to-obtain high-quality data, it is difficult for general-purpose LLMs — trained primarily on open internet data — to acquire the necessary expertise in domain-specific applications. For example, tasks in finance, healthcare, or scientific research often require expertise that is not well represented in public datasets. On the other hand, privacy-sensitive applications, such as medical diagnosis or legal document analysis, often prohibit data uploading, necessitating local deployment of models. The computational demands of LLMs inference and tuning create prohibitive infrastructure costs, posing challenges for deployment in many resource-constrained real-world scenarios. Due to these limitations, more and more specialized small language models (SLMs) are being trained for specific domains~\citep{wang2024comprehensive,Li:Zhang2024,Abdin:Aneja2024}.

The \emph{learnware} paradigm \citep{FutureOfML2016,small_model_do_big}, first proposed in 2016, offers a promising solution to these problems. A \emph{learnware dock system} accommodates numerous trained models in a unified way, the trained models are contributed by diverse developers.
Each model is associated with a \emph{specification}. Learnware = Model + Specification. The specification is generated at the model developer's side, and during the whole process the system does not touch the raw data of the developers.
The future user can submit her task request, without exposing their raw data to the system; the system will identify some helpful learnwares or even assemble some learnwares to return to user to help addressing her task. During the whole process, the learnware dock system does not touch user's raw data. The specification plays a vital role in the whole process. We will explain how the specification is generated in Section~\ref{sec:method-model}.

The learnware paradigm has been shown helpful in some common scenarios involving small machine learning models~\citep{small_model_do_big,Beimingwu2024}.
In the context of language models, we claim that this paradigm is also compelling. For example, while many developers are willing to share/sell well-trained models, they are generally reluctant to share their high-quality or proprietary data, especially when data privacy is a concern. As a result, a repository of specialized models can collectively integrate more decentralized expertise from private data sources, and may provide stronger capabilities in specialized scenarios than single large model trained solely on open-source data. Furthermore, due to the risk of catastrophic forgetting and the pursuit of general-purpose scenario, even state-of-the-art LLMs—though powerful on open-ended or generic tasks—are not convenient to show optimal performance on highly specialized applications.

As the number of specialized language models grows, a fundamental challenge is to effectively and efficiently identify and utilize helpful models for a new task in a privacy-preserving manner.  
Exhaustively loading and evaluating every candidate model is computationally infeasible, and users are generally unwilling or unable to upload private data for model identification or evaluation. 
Therefore, it is crucial to develop efficient and privacy-preserving methods for matching user requirements with the helpful expert models. This fundamental challenge is systematically considered in the learnware paradigm: the specification in each learnware uniformly characterizes model capabilities, and is the key to enable the system to function as an evolving ecosystem of reusable components: users solve tasks by identifying relevant models through specification matching rather than exposing original data, while eliminating the requirement for prohibitively costly per-model evaluations.

In this paper, we presents a preliminary attempt to apply the learnware paradigm to language models, and demonstrate the potential of the learnware paradigm to organize and utilize specialized SLMs effectively.
We simulated a learnware dock system containing about a hundred task-specific 8B-scale small language models from three domains: finance, healthcare, and mathematics. Experiments show that by simply identifying a suitable SLM learnware based on specifications for each task to perform inference, the system outperforms the base SLMs on all benchmarks. Compared to large language models, the system outperforms well-known LLMs with over 70B parameters like Qwen2.5-72B, and Llama3.1-70B-Instruct by at least 14\% in finance domain tasks. Additionally, it surpasses Flan-PaLM-540B (ranked 7th on the Open Medical LLM Leaderboard\footnote{\url{https://huggingface.co/spaces/openlifescienceai/open_medical_llm_leaderboard}}) in medical domain tasks.
We further discussed the underlying mechanism behind its advantages.

\section{How to construct specification for LLMs}
\label{sec:method-model}

In this section, we introduce how the developers, who want to submit their models to the system, construct specifications for LLMs. We describe the workflow that developers follow, from training models to submitting learnware to the learnware dock system. Additionally, we discuss specific implementations of the specifications, including how they characterize model capabilities and how they are compressed into a low-rank space to ensure computational and storage efficiency.

\begin{figure}[t]
    \centering  
    \subfloat[Workflow for Developers Uploading Learnware to the Learnware Dock System]{
        \includegraphics[width=\linewidth]{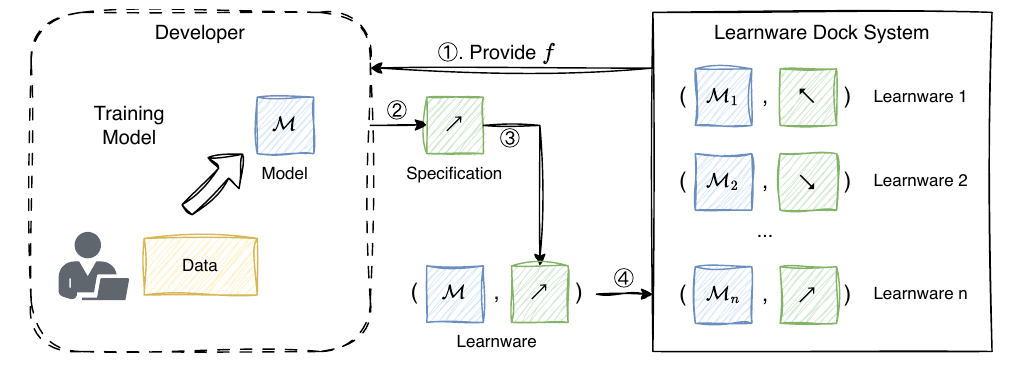}
        \label{fig:LLMSpec}
    }
    \vspace{0.01em}
    \subfloat[Workflow for Users Acquiring Learnware from the Learnware Dock System]{
        \includegraphics[width=\linewidth]{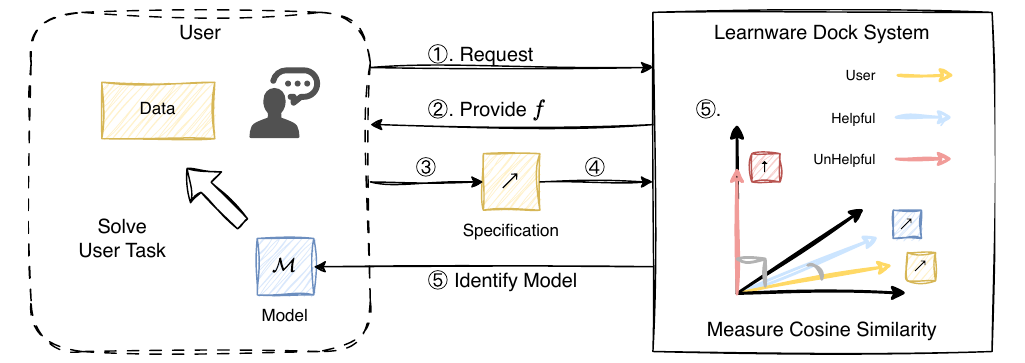}
        \label{fig:UserSpec}
    }
    \vspace{0.6em}
    \caption{Workflows of the Learnware Paradigm.
             (a) \textbf{Workflow for Developers Uploading Learnware to the Learnware Dock System:} \ding{172} LDS provides developers with the function $f$ for generating specifications. \ding{173} Developers generate a specification based on a model and data locally. \ding{174} The specification and model are combined into a learnware. \ding{175} The learnware is then submitted to the learnware dock system. (b) \textbf{Workflow for Users Acquiring Learnware from the Learnware Dock System:} \ding{172} User requests the acquisition of Learnware from LDS. \ding{173} LDS supplies the user with function $f$ for specification generation. \ding{174} User generates a specification based on her data to represent her task requirements. \ding{175} The requirement specification is submitted to the LDS. \ding{176} LDS identifies and returns relevant learnware to the user based on the cosine similarity between user specifications and learnware specifications.}
    \label{fig:lds_workflows_combined}
\end{figure}

In order to assist developers in seamlessly submitting models trained on private data to the learnware dock system (LDS), we provide a structured submission workflow. As illustrated in \Cref{fig:LLMSpec}, when a developer submits a model, LDS first supplies a dedicated function $f$. Using this function, the developer can locally generate a specification based on the model and its training data. The model and its specification are then packaged into a learnware and submitted to the LDS. This streamlined approach ensures that private training data is handled securely throughout the submission process. For common machine learning models, the function can be simply a kernel function~\citep{small_model_do_big}. For large language models, a more powerful function is exploited.

In detail, the capability of a model $h$ can be mainly represented by the conditional distribution $p(h(\mathbf{x})|\mathbf{x})$, which reflects both task semantics and model quality (alignment of $h(\mathbf{x})$ with true label $y$), and the marginal distribution of input where the model excels. By fine-tuning an extra SLM to fit the distribution, the changes in the model’s parameters, can be utilized as specification to represent the model capabilities. Roughly speaking, we use an extra SLM, say $F$, to realize the $f$ in \Cref{fig:LLMSpec}, and each language model can be characterized/distinguished by the parameter changes vector of $F$ in its functional space, which is used as the specification associated with the language model. The algorithm of building the specification for developers is presented as \Cref{alg:dev} in \Cref{apx:algs}. 

In practice, the parameter space of modern pre-trained models is typically in the hundreds of millions, making direct construction and comparison of parameter vectors derived from the weight space impractical.
Therefore, we approximate parameter vectors in a LoRA \citep{LoRA2022} manner that $\boldsymbol{\tau}\approx\tilde{\boldsymbol{\tau}}\triangleq\mathbf{BA}$ and change the weights to be optimized from $\boldsymbol{\tau}$ to $\mathbf{B}$, reducing size to less than 1\% of the pre-trained model's weights—typically under 1M parameters—greatly improving efficiency in construction, comparison, and storage. The algorithm of specification generation in the low-rank space is presented as \Cref{alg:lora} in \Cref{apx:algs}.

\section{How to serve users}
\label{sec:method-user}

The overall workflow is illustrated in \Cref{fig:UserSpec}:  user initiates a request and receives a function $f$ tailored to her needs from LDS. Next, she can locally generate a specification based on her own task data and submit it to LDS. The system then examines available learnware by comparing specification vectors, selecting those nearby the user’s specification. Finally, the user employs the returned learnware model to address her task. 

The requirements of a user task can mainly be represented by the joint distribution $p_u(\mathbf{x},y)$, reflecting the required capability to solve the user task. User can also build the task specification at her side using the SLM function $f$ received from the learnware dock system, to characterize her task requirements by the parameter changes vector without exposing her raw data to the LDS. The detailed algorithm is presented as \Cref{alg:user} in \Cref{apx:algs}. 
Then, the alignment between the model's capability and the user task requirements can be assessed by the cosine distance between parameter vector specifications.
In practice, we identify the one with the smallest cosine distance. 

Following the approximation method discussed in \Cref{sec:method-model}, we can approximate $\cos(\tilde{\boldsymbol{\tau}}_1,\,\tilde{\boldsymbol{\tau}}_2)$ by computing $\cos(\mathbf{B}_1,\,\mathbf{B}_2)$ only, because the matrix $\mathbf{A}$ is initialized randomly and does not update in the process of specification generation, thus contains less useful information. 
In this way, we can also reduce the computational overhead for computing cosine similarity.

\section{Results}

In this section, we present the experimental results of learnware paradigm on language model scenarios. Based on Beimingwu~\citep{Beimingwu2024}, 
we build a learnware dock system containing specialized SLMs from three domains: finance, healthcare, and mathematics. 
These SLMs are models fine-tuned on several specific datasets in each domain.
By selecting one suitable SLM learnware for each task-specific inference, the learnware dock system outperforms the base SLMs on all benchmark, and achieves performance comparable to a single large model on public evaluation benchmarks while requiring significantly less costs for inference. 

It is worth mentioning that as pointed in~\citet{van2024survey}, the definitions of “small” and “large” are a function of both context and time. In this paper, we follow the generalized definition of small language models (SLMs) in~\citet{wang2024comprehensive}: \emph{Given specific tasks and resource constraints, we define SLMs as falling within a range where the lower bound is the minimum size at which the model exhibits emergent abilities for a specialized task, and the upper bound is the largest size manageable within limited resource conditions.}  This definition integrates various perspectives and addresses factors related to mobile computing and capability thresholds. Here compared to 70B level LLMs, we refer 8B-scale models in our experiments as SLMs.

\subsection{Public benchmark performance}

We evaluate the learnware dock system of SLMs with under 8B parameters on public benchmarks across three domains: finance (FinBen \citep{xie2024finben}), healthcare (Open Medical LLM Leaderboard \citep{OpenMedicalLLMLeaderboard,singhal2022large}), and mathematics (part of MathEval\footnote{\url{https://matheval.ai/}} and DeepSeekMath evaluation benchmark \citep{shao2024deepseekmath}).
In the finance domain, our system outperforms 
well-known open-source large language models with at least a 14\% boost, including Llama-3.1-70B-Instruct \citep{llama3}, Qwen2.5-72B \citep{yang2024qwen2.5}, and Qwen1.5-110B \citep{qwen1.5}. In the healthcare domain, it surpasses Flan-PaLM-540B \citep{chung2024scaling} and in the mathematics domain, it significantly outperforms baseline methods such as random model selection. 

\subsubsection{Settings}
\label{sec:app-settings}

To construct a diverse collection of fine-tuned 8B-level language models, we first train multiple models under different configurations to simulate the models submitted by many different developers. Details on fine-tuning, the construction of model hub containing task-specific SLMs, and evaluation benchmarks for each scenario are provided in \Cref{sec:setting}.
Following the pipeline described in \Cref{sec:method-model,sec:method-user}, we generate specifications for each model, and select the most suitable learnware for user task requirements. The identified learnware is then evaluated on the corresponding task to assess overall system performance under the \textit{Task-Level} evaluation setting. 
Task-Level evaluation means that the user can generate her requirement specification from the whole task data.
Furthermore, we compare learnware dock system against several contenders, including:
\begin{itemize}[left=0pt]
    \item Ways to utilize specialized SLMs. This contains a baseline algorithm, \texttt{Random} learnware selection, and two oracle-style strategies with access to the full evaluation results of all candidate models, the \texttt{Best-single} model and \texttt{Oracle}. \texttt{Best-single} refers to the model with the highest average score among the learnware candidates, and \texttt{Oracle} is the optimal performance of utilizing the candidate SLM learnwares by choosing one model for one task, which selects the ground-truth best performing model on each user task.
    \item Base models used for fine-tuning.
    \item Well-known general-purpose LLMs with over 70B parameters.
\end{itemize}
For simplicity, considering that each specialized SLM in the experiments is obtained by fine-tuning and performs well on specific datasets, the learnware specifications in our experiments is generated by fitting $p(y|\mathbf{x})$, thus represent the model capability by capturing the characteristics of fine-tuning data. Additionally, our specification implementation is available in the \texttt{learnware} package, and detailed hyperparameter configurations can be found in \Cref{sec:setting-overview}.

\subsubsection{Finance}

\begin{figure}[t]
    \centering
    \includegraphics[width=\linewidth]{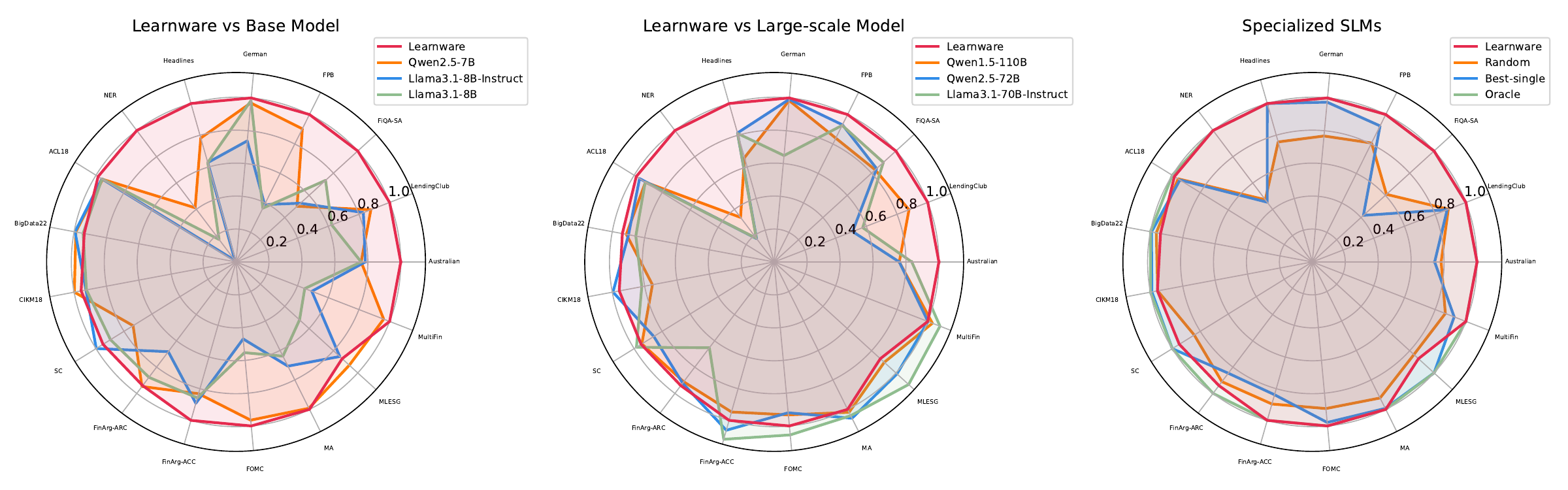}
    \caption{\textbf{Performances on financial LLM evaluation benchmark.} The performance metrics are also normalized relative to \texttt{Oracle}. Detailed performance values are shown in \Cref{tab:finance}.}
    \label{fig:llm-finance}
\end{figure}

\begin{table}[t]
\setlength{\tabcolsep}{1pt}
\caption{Performance value of different methods or language models on our financial evaluation benchmark. Full score is 100. The highest score among all contenders, except oracle-style strategies \texttt{Best-Single} and \texttt{Oracle}, is emphasized in bold. ``*'' denotes the best score of all contenders in ``Specialized SLMs'' except \texttt{Oracle}.}
\label{tab:finance}
\centering
\resizebox{\textwidth}{!}{
\begin{tabular}{ll|cccccc|cc|cc}
\toprule
\multicolumn{2}{c}{User Task} & \multicolumn{3}{c}{Base Models}             & \multicolumn{3}{c}{Large-scale Models}       & \multicolumn{4}{c}{Specialized SLMs}\\
\cmidrule(lr){1-2}\cmidrule(lr){3-5}\cmidrule(lr){6-8}\cmidrule(lr){9-12}
Dataset & Metric            & Qwen2.5-7B     & Llama3.1-8B-Instruct & Llama3.1-8B & Qwen1.5-110B & Qwen2.5-72B    & Llama3.1-70B-Instruct & Random & \textbf{Learnware} & Best-single & Oracle  \\ \midrule
Australian & Acc & 43.17          & 44.60                 & 43.17        & 43.17        & 43.17          & 47.48                  & 44.45       & \textbf{56.83*}            & 42.21 & 56.83      \\
LendingClub & Acc & 80.82          & 76.33                 & 57.34        & 80.82        & 47.01          & 53.07                  & 81.52       & \textbf{92.07*}            & 80.82 & 92.07      \\
FiQA-SA & Acc    & 38.30          & 40.43                 & 56.17        & 63.40        & 64.26          & 68.51                  & 46.53       & \textbf{76.38*}            & 32.06 & 76.38      \\
FPB  & Acc       & 76.08          & 32.78                 & 30.72        & 70.72        & 78.35          & 78.04                  & 67.95       & \textbf{84.25*}            & 77.73 & 84.25      \\
German & Acc     & 65.00          & 49.50                 & 66.00        & 66.00        & 66.50          & 43.50                  & 51.50       & \textbf{67.06*}            & 65.33 & 67.06      \\
Headlines & AvgF1  & 74.81          & 59.95                 & 59.95        & 62.96        & 77.84          & 77.53                  & 72.43       & \textbf{95.61*}            & 95.61* & 95.61      \\
NER   & EntityF1      & 21.75          & 0.62                  & 9.01         & 17.89        & 9.36           & 9.52                   & 24.99       & \textbf{52.79*}            & 23.98 & 52.79      \\
ACL18 & Acc      & 51.10          & 51.40                 & 51.34        & 49.30        & 51.56          & 49.38                  & 51.42       & \textbf{52.82*}            & 50.71 & 53.63      \\
BigData22 & Acc  & 55.30          & \textbf{55.57}        & 52.79        & 51.02        & 50.27          & 47.76                  & 53.86      & 52.40            & 55.52*  & 55.88               \\
CIKM18  & Acc    & \textbf{58.44} & 54.24                 & 54.07        & 44.01        & 58.27          & 47.86                  & 55.89       & 55.99            & 57.98*  & 58.52              \\
SC  & Acc        & 65.14          & 88.48                 & 79.45        & 83.75        & 76.17          & \textbf{87.16}                  & 74.71       & 84.17 & 88.61*   & 88.61               \\
FinArg-ARC & Acc & \textbf{64.78} & 46.67                 & 60.00        & 62.32        & 63.04          & 44.64                  & 62.27       & 64.31*            &  57.87 & 68.36              \\
FinArg-ACC & Acc & 48.30          & 51.81                 & 49.85        & 55.01        & 61.71          & \textbf{65.02}         & 52.08       &  58.08*           & 48.68  & 58.08              \\
FOMC   & Acc     & 60.48          & 29.44                 & 34.68        & 58.47        & 57.66          & \textbf{66.13}         & 56.05       & 62.70*          & 61.36  & 62.70               \\
MA     & Acc     & 79.20          & 56.40                 & 51.00        & 81.40        & \textbf{84.60} & 83.20                  & 73.64       & 79.81*            & 79.27  & 79.81              \\
MLESG  & Acc    & 35.67          & 32.67                 & 20.00        & 34.67        & 38.67          & \textbf{42.33}         & 31.99       & 33.42          & 38.33*  & 38.33               \\
MultiFin & Acc   & 60.99          & 31.32                 & 28.39        & 65.38        & 63.55          & \textbf{68.50}         & 54.96       & 63.46*            & 58.61  & 63.46              \\
\midrule
Avg.(↑) &   & 57.61          & 47.19                 & 47.29        & 58.25        & 58.35          & 57.63                  & 56.25  & \textbf{66.60*}          & 59.69 & 67.79 \\ 
Avg. rank(↓) & & 5.94 & 7.35 & 7.82 & 5.94 & 4.71 & 5.24 & 6.47 & \textbf{2.88*} & 5.47 & - \\
\textbf{Learnware} (win/tie/loss) & & 13/0/4 & 15/0/2 & 16/0/1 & 14/0/3 & 12/0/5 & 11/0/6 & 16/0/1 & - & 12/1/4 & 0/11/6 \\
Oracle (win/tie/loss) & & 17/0/0 & 17/0/0 & 17/0/0 & 15/0/2 & 13/0/4 & 12/0/5 & 17/0/0 & 6/11/0 & 14/3/0 & - \\
\bottomrule
\end{tabular}
}
\end{table}

Our system, surprisingly, demonstrates strong performance across financial tasks, achieving the highest average score among all contenders, delivering an nearly 14\% improvement compared with the best large-scale model Qwen2.5-72B, showed in \Cref{fig:llm-finance} and \Cref{tab:finance}. It ranks first among strategies utilizing task-specific SLMs except \texttt{Oracle} in 13 out of 17 tasks, identifies the optimal learnware (tied with \texttt{Oracle}) on 11 and outperforms all contenders in 8. For tasks whose training split in FIT \citep{xie2023pixiu} for fine-tuning, the system performs well due to the availability of specialized models in the learnware candidates. However, in the three Forecasting tasks (BigData22, ACL18, CIKM18), fine-tuned models often underperform compared to their base versions, likely due to the raw tabular nature of the data, a challenge also noted in prior research \citep{xie2023pixiu}. Besides, our system outperforms the \texttt{Single-Combined} model, which is fine-tuned using a combination of 10 datasets in FIT, shown in \Cref{tab:fin-extra}. Details about the \texttt{Single-Combined} model is introduced in \Cref{sec:fin-extra}. 

These results significantly verify the potential of utilizing many specialized learnwares to solve user tasks and highlight two key insights: \textbf{(1)} Our system can match or surpass large-scale models with over 70B parameters under the \textit{Task-Level} evaluation setting, while requiring only the memory for models under 8B efficiently. \textbf{(2)} Managing a diverse set of specialized models through our system can achieve better results than dedicating extensive resources to the fine-tuning of a single model with the same size, known as the \texttt{Single-Combined} model, suggested by results in \Cref{tab:fin-extra}. 

\subsubsection{Healthcare}

\begin{table}[t]
\centering
\caption{Performance value of different methods or language models on our medical evaluation benchmark. Full score is 100. The highest score among all contenders, except oracle-style strategies \texttt{Best-Single} and \texttt{Oracle}, is emphasized in bold. ``*'' denotes the best score of all contenders in ``Specialized SLMs'' except \texttt{Oracle}.}
\label{tab:med}
\resizebox{\textwidth}{!}{
\begin{tabular}{l|cc|cc|cc}
\toprule
\multicolumn{1}{c}{User Task} & \multicolumn{1}{c}{Base Model} & \multicolumn{1}{c}{Large-scale Model} & \multicolumn{4}{c}{Specialized SLMs} \\
\cmidrule(lr){1-1} \cmidrule(lr){2-2} \cmidrule(lr){3-3} \cmidrule(lr){4-7}
Dataset                   & Qwen2.5-7B          & Flan-PaLM-540B             & Random      & \textbf{Learnware}    & Best-single & Oracle         \\
\midrule
medmcqa                & 59.93               & 57.60                      & 60.20       & \textbf{62.49*} & 62.49* & 62.49      \\
medqa\_4options        & 64.18               & \textbf{67.60}             & 63.74       & 65.59*          & 64.81  & 65.59              \\
anatomy                & \textbf{71.85}               & 63.70                      & 71.33       & \textbf{71.85*}          & 70.37 & 72.96       \\
clinical\_knowledge    & 77.36               & \textbf{80.40}             & 78.21       &   78.87*        & 78.49 & 79.25               \\
college\_biology       & 82.64               & \textbf{88.90}             & 84.34       &   85.42*        & 84.03  & 86.11              \\
college\_medicine      & 69.36               & \textbf{76.30}             & 69.02       &   69.36*        & 68.79  & 69.95              \\
medical\_genetics      & \textbf{87.00}               & 75.00                      & 86.95       & \textbf{87.00} & 89.00*  & 89.00              \\
professional\_medicine & 78.68               & \textbf{83.80}             & 77.37       & 79.78*          & 78.68  & 79.78              \\
pubmedqa               & 75.20               & \textbf{79.00}             & 75.67       &    75.80       & 76.80*  & 76.80              \\
\midrule
Avg.                    & 74.02         & 74.70                & 74.09     &   \textbf{75.13*}        & 74.83 & 75.77 \\
Avg. rank(↓) & 4.44 & 2.67 & 4.89 & \textbf{2.56*} & 3.56 & - \\
\textbf{Learnware} (win/tie/loss) & 6/3/0 & 3/0/6 & 9/0/0 & - & 6/1/2 & 0/3/6 \\
Oracle (win/tie/loss) & 9/0/0 & 3/0/6 & 9/0/0 & 6/3/0 & 6/3/0 & - \\
\bottomrule
\end{tabular}
}
\end{table}

The performance of all methods is shown in \Cref{fig:llm-med} and \Cref{tab:med} in \Cref{sec:extra}. Our system achieves the highest average score across 9 tasks, even surpassing the large-scale model Flan-PaLM-540B. This demonstrates that by leveraging multiple models with fewer than 8B parameters, our system can outperform a single large-scale model in task-specific scenarios. Among SLM utilization strategies, \texttt{Learnware} performs best in 7 out of 9 tasks, tied with \texttt{Oracle} in 6.
Furthermore, the fact that our system surpasses \texttt{Best-single} highlights that its effectiveness comes not from a single exceptionally strong model but from its specification design, identification mechanism and the collective strength of all candidate models.

\subsubsection{Mathematics}

The results are presented in \Cref{fig:llm-math} and \Cref{tab:math} in \Cref{sec:extra}. Our system achieves optimal identification performance (tied with \texttt{Oracle}) in 10 out of 16 tasks and even outperforms all other contenders except oracle-style strategies \texttt{Best-Single} and \texttt{Oracle} in 5. However, the large-scale model achieves the highest average score and even beats \texttt{Oracle} (which denotes the optimal performance using one of our 8B-level models). This is likely due to their strong mathematical reasoning abilities that lack in smaller models, rather than a shortcoming of our method, as evidenced by the minimal difference in the "win/tie/loss" of \texttt{Learnware} and \texttt{Oracle} on Qwen1.5-110B.

\subsection{Deployment on the Beimingwu platform}

\textbf{Beimingwu.} \texttt{Beimingwu} platform \citep{Beimingwu2024} is the first open-source learnware research platform which supports the entire process of learnware paradigm, including the submitting, usability testing, organization, identification, deployment and reuse of learnwares. The system lays the foundation for research experiments and demos in learnware related algorithms and systems.

\textbf{Implementations.} Our method has been integrated into \texttt{Beimingwu} platform to facilitate SLM learnware identification in open, real-world environments. 
Specifically, we implement the class \texttt{GenerativeModelSpecification} for constructing parameter vector specifications with Qwen2.5-0.5B as the pre-trained model, which is also part of the learnware package\footnote{Github Repository: \url{https://github.com/Learnware-LAMDA/Learnware}}. The code for reproducing the experimental results in this paper is also provided in the Github Repository.

\section{Why does it work}

While the empirical results demonstrate the effectiveness of our approach, the underlying mechanisms enabling this capability merit further investigation. In this section, we try to explain the key factors that contribute to the strength of the Learnware Dock System (LDS) in specialized scenarios.

\begin{figure}[t!]
    \centering
    \includegraphics[width=\linewidth]{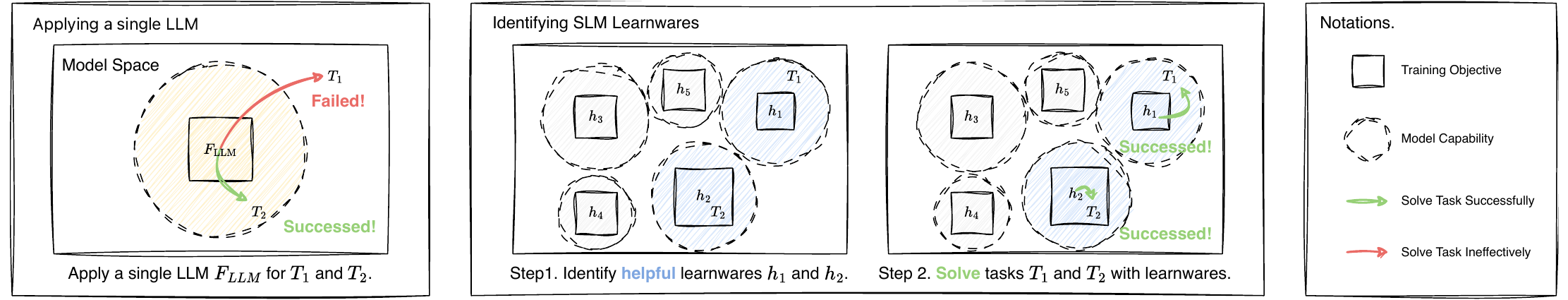}
    \caption{A diverse collection of SLM learnwares, though each of them falls behind LLMs for general unseen tasks, can surpass the capability boundaries of LLMs in specialized scenarios. For example, for the user task $T_1$ which can not be solved by LLM, the learnware $h_1$, though not originally designed for $T_1$, can handle it.
    }
    \label{fig:beyound}
\end{figure}

{\bf Accessing and Integrating Decentralized Expertise}. 
AI applications exhibit a long-tail distribution of specialized scenarios, countless specific tasks are individually rare but collectively crucial. Monolithic LLMs are typically trained on centralized, publicly available datasets, and it inherently limits their access to the vast amount of valuable, specialized, or proprietary expertise within private datasets across various organizations and individuals.
In contrast, the learnware dock system overcomes this limitation by aggregating models from developers who share well-trained/post-trained SLMs without revealing their private or high-quality data. This allows the system to access a wider range of domain-specific expertise that LLMs may miss. While in-context learning can improve how LLMs use existing capabilities, it cannot fill fundamental gaps in their pre-trained capabilities base. Our results in finance highlight this advantage, as the learnwares in the system leverage specialized expertise in contributed models, surpassing the capability boundaries of even LLMs with over 70B parameters. As shown in \Cref{fig:beyound}, though individual SLMs have narrower capability ranges than LLMs, the collective LDS enables access to expertise beyond public data boundaries, where individual learnwares address specific tasks while the system maintains broad applicability.

Besides, specialized SLMs in our experiments, while smaller in total parameter count, are built upon capable pre-trained foundations (like 8B-scale base models) and undergo subsequent fine-tuning phases highly focused on a specific domain/task. This focused specialization allows the SLM's relatively smaller parameter space to be efficiently dedicated to mastering the crucial patterns essential for high performance on that specific task. In contrast, in pursuit of generality, LLMs often face challenges like negative transfer, where optimization for one task can degrade performance on another. This may be caused by task conflicts where optimization for different objectives contradict, e.g., sentiment analysis requiring subjectivity versus fact-based tasks needing objectivity; gradient interference from conflicting parameter update directions across tasks; and resource competition where limited capacity is split across tasks, weakening performance on specific ones. 
Our system circumvents this by maintaining independently post-trained models that concentrate on specific task requirements and domains. 

A core technical challenge in leveraging a large library of specialized models is efficiently identifying the most helpful ones for a given user task without exposing original data. The specification framework of the learnware paradigm provides the crucial mechanism for addressing this. Users solve tasks by identifying relevant models through specification matching rather than exposing original data, eliminating the requirement for prohibitively costly per-model evaluations. By making efficient and privacy-preserving learnware identification possible, specification enables the learnware dock system to be open and collaborative such that the LDS can possess the above advantages. 

\section{Related work}
\label{sec:related-work}

The learnware paradigm \citep{FutureOfML2016, small_model_do_big} offers a systematic way to manage well-trained models and leverage their capabilities to assist users in solving their tasks, rather than training from scratch. 
The Reduced Kernel Mean Embedding (RKME) \citep{small_model_do_big,ExplicitLabelExploitationZhou2024} is proposed to identify useful learnwares by matching their original data distribution with the user’s data distribution, and its unsupervised version has been demonstrated effective in various tasks~\citep{WuXiZhu,Beimingwu2024}. Based on the RKME specification, there have been advances on 
efficient identification \citep{Evolvable2024,XieYi2023}, extension to heterogeneous feature spaces \citep{HeterogeneousWithoutAuxiliary2023,tan2024towards}, heterogeneous label spaces \citep{HeterogeneousLabel2023} and unseen tasks \citep{UnseenJobs2021}. These explorations verify the feasibility of the learnware paradigm. Furthermore, the privacy-preserving ability of the learnware paradigm for developers' raw data has been theoretically proved~\citep{Privacy2024}. 
To facilitate research on learnware, Beimingwu\footnote{Beimingwu Homepage: \url{https://bmwu.cloud}} learnware research platform has been released \citep{Beimingwu2024}.
For common machine learning models, RKME specification can be utilized, and parameter vector specification in this paper focuses on characterizing deep learning models built on high-dimensional and unstructured data.

Different from all existing approaches trying to select appropriate LLM model for specific user needs~\citep{ong2024routellm,frick2025prompt,aggarwal2024automix,Jiang0L23}, learnware exhibits a fundamentally new way. 
First, learnware does not rely on routing models built on auxiliary external datasets, which are hard to obtain and scale in real-world scenarios. By utilizing a pre-computed specification characterizing model capabilities and user requirements, the learnware paradigm is naturally scalable to ever-increasing number of models, without re-training the routing models constantly with new model being submitted. Besides, the learnware paradigm does not expose the raw data of users during model identification process, instead utilizing efficient specification matching, enabling privacy-preserving model identification. At the same time, instead of utilizing LLMs as black-box API, by characterizing the internal functional capability and bias of models, learnware specifications represent their unique generalization and reusability properties on new tasks, and facilitate deep understanding of model's parameter space.

\section{Conclusion}

\label{sec:conclusion-discussion}
In this paper, we demonstrate the compelling potential of the learnware paradigm for language models, particularly highlighting its value in scenarios characterized by strict constraints such as data privacy, computational cost efficiency, and the need for expert-level performance on well-defined specialized tasks. By organizing specialized SLMs with privacy-preserving parameter vector specifications, our system can achieve promising performance in specialized scenarios and significant lower inference costs. Due to decentralized expertise integration, focused specialization and specification framework for efficient and privacy-preserving model identification, learnware paradigm shows a promising way for handling numerous specialized scenarios in real world by establishing an open and collaborative learnware dock system. 

While our paper focused on enabling users to acquire these specialized models to solve a specific task by local deployment, in the future, these specialized learnwares are also crucial for empowering general AI systems and agents by providing a valuable source of expert capabilities that can be identified with specifications and leveraged for complex and specialized sub-tasks. Besides, once the system accumulates a sufficient collection of learnwares and develops a deeper understanding of the specification space, it could allow developers to upload models without requiring local specification generation. In this mode, the system would analyze the new model's characteristics and behavior, potentially through interaction with existing learnwares~\citep{Beimingwu2024,Evolvable2024}, and generate its specification internally. This process would significantly lower the contribution barrier for developers while ensuring that these models can be accurately and efficiently identified by users through specification matching, further enriching the learnware ecosystem.

\textbf{Limitations.} Despite the great potential, our work makes initial exploration on LLM learnware identification.
Due to the nature of supervised fine-tuning in enhancing task-specific expertise, the parameter vector specification currently serves as an initial attempt for identifying task-specific capabilities needed to solve user tasks.
However, LLMs possess more advanced capabilities beyond this, which have not been characterized in our experiments.
Actually, other post-training techniques such as RLHF alignment~\citep{ouyang2022training}, also ultimately affect the model's parameters, allowing the parameter vector specification to capture the capabilities learned by the model.
Additionally, it would be valuable to explore more efficient methods for generating the parameter vector specification, reducing the need to load the full pre-trained model locally~\citep{Side-Tuning2024} to enhance the feasibility in resource-constrained environments.

\section*{Acknowledgements}

This work was supported by the National Science Foundation of China (Grant No. 62250069). We thank Xiao-Chuan Zou for his valuable technical support in system development and deployment. We thank Jian-Dong Liu and Hai-Tian Liu for their help in Beimingwu implementation. We thank Tian-Zuo Wang, Han-Jia Ye, Lan-Zhe Guo, Jia-Wei Shan and Hao-Yi Lei for their helpful discussions. We appreciate the Polixir team for their support on the system.

\bibliographystyle{plainnat}
\bibliography{task-spec}

%%%%%%%%%%%%%%%%%%%%%%%%%%%%%%%%%%%%%%%%%%%%%%%%%%%%%%%%%%%%

\newpage
\appendix
\renewcommand{\thetable}{A\arabic{table}}
\renewcommand{\thefigure}{A\arabic{figure}}

\section{Algorithm details}
\label{apx:algs}

In this section, we give the concrete implementations of the algorithms for constructing the parameter vector specification for both developers and users. For learnwares, The algorithm for generating model vectors to sketching the model capabilities is presented as \Cref{alg:dev}. 

\begin{algorithm}[ht]
\caption{Build the parameter vector specification for developers}\label{alg:dev}
\begin{algorithmic}[1]
\Require The model $h(\mathbf{x})$ trained on the task $(\mathcal{D}, \mathcal{L})$.
\State initialize pre-trained model $f(\mathbf{x},\,\theta_0+\boldsymbol{\tau})$ with an all-0 task vector $\boldsymbol{\tau}$
\State construct function $g$ such that $g\circ f(\mathbf{x},\,\theta_0+\boldsymbol{\tau})\in \mathcal{Y}$ if necessary
\Repeat{}
    \For{each mini-batch $\mathcal{B} \subset \mathcal{D}$}
    \Comment{$\mathcal{B}=\{(\mathbf{x}_i,\,y_i)\}$}
        \State 
        $\hat{y}_i\gets h(\mathbf{x}_i)$ with no\_grad\,() for $\mathbf{x}_i$ in $\mathcal{B}$ \label{line:y-hat}
        \State $y_i'\gets g\circ f(\mathbf{x}_i,\,\theta_0+\boldsymbol{\tau})$ for $\mathbf{x}_i$ in $\mathcal{B}$ 
        \State update model vector: $\boldsymbol{\tau} \gets \boldsymbol{\tau} - \eta \sum_{i} \nabla_{\boldsymbol{\tau}} \mathcal{L}(y_i',\, \hat{y}_i)$\Comment{Fitting $p(\hat{y}|\mathbf{x})$ through $\mathcal{L}$.} \label{line:update}
    \EndFor
\Until{converged}
\State \Return $\boldsymbol{\tau}$
\end{algorithmic}
\end{algorithm}

For user-generated task vectors, the model prediction $\hat{y}$ should be replaced with $y$
throughout, 
\begin{equation}
    \boldsymbol{\tau}_u = \mathop{\arg\min}_{\boldsymbol{\tau}}\sum_{(\mathbf{x},y)\in \mathcal{D}_u} \mathcal{L}(g\circ f(\mathbf{x},\,\theta_0+\boldsymbol{\tau}),\,y),
\end{equation}
and \Cref{line:y-hat} in \Cref{alg:dev} can be omitted, shown in \Cref{alg:user}. It means that we are sketching the required capability $p_u(y|\mathbf{x})$ for the user task $(\mathcal{L}_u,\,\mathcal{D}_u)$. 

\begin{algorithm}[ht]
\caption{Build the parameter vector specification for users}\label{alg:user}
\begin{algorithmic}[1]
\Require The user task $(\mathcal{D}_u, \mathcal{L}_u)$ required to solve.

\State initialize pre-trained model $f(\mathbf{x},\,\theta_0+\boldsymbol{\tau})$ with an all-0 task vector $\boldsymbol{\tau}$
\State construct function $g$ such that $g\circ f(\mathbf{x},\,\theta_0+\boldsymbol{\tau})\in \mathcal{Y}$ if necessary
\Repeat{}
    \For{each mini-batch $\mathcal{B} \subset \mathcal{D}_u$}
    \Comment{$\mathcal{B}=\{(\mathbf{x}_i,\,y_i)\}$}
        \State $y_i'\gets g\circ f(\mathbf{x}_i,\,\theta_0+\boldsymbol{\tau})$ for $\mathbf{x}_i$ in $\mathcal{B}$
        \State update task vector: $\boldsymbol{\tau} \gets \boldsymbol{\tau} - \eta \sum_{i} \nabla_{\boldsymbol{\tau}} \mathcal{L}(y_i',\, y_i)$\Comment{Fitting $p_u(y|\mathbf{x})$ through $\mathcal{L}$.}
    \EndFor
\Until{converged}
\State \Return $\boldsymbol{\tau}$
\end{algorithmic}
\end{algorithm}

Besides, \Cref{alg:lora} shows the specification generation algorithm with approximation of parameter vectors in the low rank space.

\begin{algorithm}[ht]
\caption{Build the parameter vector specification in the low-rank space}\label{alg:lora}
\begin{algorithmic}[1]
\Require The model $h(\mathbf{x})$ trained on the task $(\mathcal{D}, \mathcal{L})$.
\State initialize pre-trained model $f(\mathbf{x},\,\theta_0+\mathbf{B}\mathbf{A})$ with matrices $\mathbf{A}$ and $\mathbf{B}$.
\State construct function $g$ such that $g\circ f(\mathbf{x},\,\theta_0+\mathbf{B}\mathbf{A})\in \mathcal{Y}$ if necessary
\Repeat{}
    \For{each mini-batch $\mathcal{B} \subset \mathcal{D}$}
    \Comment{$\mathcal{B}=\{(\mathbf{x}_i,\,y_i)\}$}
        \State 
        $\hat{y}_i\gets h(\mathbf{x}_i)$ with no\_grad\,() for $\mathbf{x}_i$ in $\mathcal{B}$
        \State $y_i'\gets g\circ f(\mathbf{x}_i,\,\theta_0+\mathbf{B}\mathbf{A})$ for $\mathbf{x}_i$ in $\mathcal{B}$
        \State update model vector: $\mathbf{B} \gets \mathbf{B} - \eta \sum_{i} \nabla_{\mathbf{B}} \mathcal{L}(y_i',\, \hat{y}_i)$\Comment{Only update matrix $\mathbf{B}$}
    \EndFor
\Until{converged}
\State \Return Flatten($\mathbf{B}$)
\end{algorithmic}
\end{algorithm} 

\section{Detailed experimental results}

\subsection{Detailed settings}
\label{sec:setting}
\subsubsection{Overview}
\label{sec:setting-overview}
\paragraph{Fine-tuning details.} In order to facilitate the production of a large number of models, the fine-tuning follows a similar process. We totally use three base models, Llama3.1-8B, Llama3.1-8B-Instruct and Qwen2.5-7B. We only do supervised fine-tuning (SFT) on them using LoRA \citep{LoRA2022} to gain the fine-tuned models in the model hub. The text data we use is mostly presented in Alpaca instruction format \citep{taori2023stanford}. The loss is computed only on the answer tokens. We employ a cosine learning rate decay schedule and a warm-up ratio of 0.03. The per-device batch size is set to 2. The LoRA target modules is set to all-linear. For a given instruction tuning dataset, we try multiple sets of hyperparameters. We use validation set loss to select the best checkpoint and hyperparameters. Training is conducted on one A100 node containing 4 GPUs.

\paragraph{Evaluation details.} For convenience and authority, we evaluate models on all the benchmark tasks using the EleutherAI's \texttt{lm-evaluation-harness} \citep{eval-harness}, the backend for Hugging Face's popular Open LLM Leaderboard \citep{open-llm-leaderboard-v2}. For finance and mathematics scenario, evaluation is conducted on 4 A6000 GPUs. For healthcare scenario, we use one A100 GPU to do evaluation of 8B models and 4 A100 GPUs to do evaluation of large-scale models.

\paragraph{Hyperparameter configurations.} For our method, we use the Qwen2.5-0.5B as the model for parameter vectors generation. During the training process, we use the AdamW optimizer with L2 weight decay of 0.5 and L1 weight decay of 1.0. We employ a peaking learning rate of $1 \times 10^{-5}$ following a cosine decay schedule after several steps of warm-up with warm-up ratio of 0.03. For each dataset, the training step is always 400. The per-device batch size is set to 8. The LoRA rank is set to 16, alpha is set to 32, and the target modules is set to \texttt{"q\_proj"}, \texttt{"k\_proj"} and \texttt{"v\_proj"}. Training of each parameter vector is conducted on one A100 GPU.

\paragraph{Others.} Since the specification generation for developers in our experiments is to fit $p(y|\mathbf{x})$ but not $p(h(\mathbf{x})|\mathbf{x})$, we do not distinguish between different models fine-tuned with the same instruction dataset, so if our method select a learnware for solving a given task, the performance is actually calculated by the average of all the models with the selected instruction dataset. That is, all values of methods in the ``Specialized SLMs'' column are the average performance of all models fine-tuned using the selected instruction dataset. It is worth mentioning that we do not assume available auxiliary external dataset~\citep{Zhang:Zhan:Ye2025} for LLM selection in our experiments.

\subsubsection{Finance}
\paragraph{Model hub details.} We use the part of the PIXIU's FIT \citep{xie2023pixiu} financial instruction tuning dataset to fine-tune three base models, Llama3.1-8B, Llama3.1-8B-Instruct and Qwen2.5-7B, and finally the model hub consists of 75 models. Specifically, we use 10 datasets in FIT: one Information Extraction (IE) task, Named Entity Recognition (NER); three Textual Analysis (TA) tasks, FPB, FiQA-SA, Headlines; three Risk Management (RM) tasks, German, Australian, LendingClub; three Forecasting (FO) tasks, BigData22, ACL18, CIKM18. The taxonomy follows the financial evaluation benchmark FinBen \citep{xie2024finben}, which is an advance version of PIXIU developed by the same institution. Slightly different from FIT, for each dataset we only use one instruction following the Alpaca instruction format, and we use the train split to do fine-tune training, the validation split to calculate validation set loss for selecting checkpoints and hyperparameters. We do not use the test split which is served as evaluation benchmark. All the raw dataset we use is available in Hugging Face platform \footnote{\url{https://huggingface.co/collections/TheFinAI/english-evaluation-dataset-658f515911f68f12ea193194}}.

Actually we first compare the performance of the three base models on the above datasets and try to fine-tune the base model with relatively better performance on the corresponding dataset. Therefore, finally in our model hub, there are fine-tuned Llama3.1-8B on FiQA-SA dataset, fine-tuned Llama3.1-8B-Instruct on Australian dataset, and fine-tuned Qwen2.5-7B on each of the rest. The performance of three base models and all models in model hub is shown in \Cref{tab:fin-models}.

\begin{table}[p]
\centering
% \tiny
\renewcommand{\arraystretch}{1.5}
\caption{Performance value of all models in model hub in finance scenario.}
\label{tab:fin-models}
\resizebox{\textwidth}{!}{
\begin{tabular}{@{}c|ccccccccccccccccc|c@{}}
\toprule
\multirow{2}{*}{\diagbox{Models}{Tasks}}                & Australian & LendingClub & FiQA-SA & FPB    & German & Headlines & NER        & ACL18  & BigData22 & CIKM18 & SC     & FinArg-ARC & FinArg-ACC & FOMC   & MA     & MLESG  & MultiFin & Avg          \\
                & (Acc)        & (Acc)         & (Acc)     & (Acc)    & (Acc)    & (avg\_f1)   & (entity\_f1) & (Acc)    & (Acc)       & (Acc)    & (Acc)    & (Acc)        & (Acc)        & (Acc)    & (Acc)    & (Acc)    & (Acc)      &              \\
\midrule
Qwen2.5-7B             & 43.17                          & 80.82                           & 38.30                       & 76.08                   & 65.00                      & 74.81                         & 21.75                          & 51.10                     & 55.30                         & 58.44                      & 65.14                   & 64.78                          & 48.30                          & 60.48                    & 79.20                   & 35.67                     & 60.99                        & 57.61                   \\
Llama-3.1-8B-Instruct  & 44.60                          & 76.33                           & 40.43                       & 32.78                   & 49.50                      & 59.95                         & 0.62                           & 51.40                     & 55.57                         & 54.24                      & 88.48                   & 46.67                          & 51.81                          & 29.44                    & 56.40                   & 32.67                     & 31.32                        & 47.19                   \\
Llama-3.1-8B           & 43.17                          & 57.34                           & 56.17                       & 30.72                   & 66.00                      & 59.95                         & 9.01                           & 51.34                     & 52.79                         & 54.07                      & 79.45                   & 60.00                          & 49.85                          & 34.68                    & 51.00                   & 20.00                     & 28.39                        & 47.29                   \\
\midrule
Australian-1           & 66.91                          & 74.02                           & 40.43                       & 32.37                   & 46.50                      & 59.95                         & 8.13                           & 51.18                     & 55.23                         & 55.03                      & 86.90                   & 49.28                          & 52.53                          & 30.44                    & 67.20                   & 28.67                     & 29.85                        & 49.10                   \\
Australian-2           & 46.76                          & 76.44                           & 41.70                       & 33.09                   & 49.50                      & 59.95                         & 3.45                           & 51.77                     & 53.60                         & 55.64                      & 87.36                   & 47.39                          & 52.01                          & 33.27                    & 70.00                   & 31.00                     & 30.77                        & 48.45                   \\
Australian-3           & 47.48                          & 79.64                           & 39.15                       & 34.64                   & 44.00                      & 59.95                         & 7.53                           & 51.75                     & 54.35                         & 54.77                      & 86.79                   & 43.91                          & 52.63                          & 35.69                    & 72.20                   & 32.33                     & 27.66                        & 48.50                   \\
Australian-4           & 63.31                          & 75.88                           & 51.91                       & 54.85                   & 54.00                      & 59.95                         & 23.06                          & 51.91                     & 55.30                         & 54.94                      & 84.26                   & 47.83                          & 52.73                          & 35.08                    & 67.80                   & 29.67                     & 32.60                        & 52.65                   \\
Australian-5           & 47.48                          & 79.30                           & 39.57                       & 32.27                   & 46.50                      & 59.95                         & 7.86                           & 51.32                     & 55.91                         & 55.82                      & 86.62                   & 47.39                          & 53.15                          & 31.85                    & 68.20                   & 32.00                     & 29.30                        & 48.50                   \\
Australian-6           & 53.96                          & 80.12                           & 37.45                       & 33.09                   & 41.50                      & 59.95                         & 10.15                          & 52.04                     & 55.64                         & 55.64                      & 86.91                   & 44.35                          & 52.43                          & 35.48                    & 69.60                   & 30.33                     & 32.42                        & 48.89                   \\
Australian-7           & 69.06                          & 80.42                           & 39.15                       & 35.05                   & 63.50                      & 59.95                         & 14.12                          & 50.78                     & 55.57                         & 54.07                      & 86.40                   & 47.54                          & 52.53                          & 34.48                    & 70.80                   & 29.00                     & 28.94                        & 51.26                   \\
Australian-8           & 59.71                          & 80.34                           & 46.81                       & 42.78                   & 66.00                      & 59.95                         & 17.88                          & 51.40                     & 54.62                         & 53.98                      & 86.93                   & 44.78                          & 52.22                          & 34.68                    & 83.80                   & 29.67                     & 31.32                        & 52.76                   \\
LendingClub-1          & 43.17                          & 80.82                           & 45.11                       & 74.85                   & 59.00                      & 72.19                         & 29.23                          & 51.02                     & 55.16                         & 58.53                      & 71.27                   & 65.36                          & 53.87                          & 61.09                    & 69.60                   & 30.00                     & 61.17                        & 57.73                   \\
LendingClub-2          & 43.17                          & 97.51                           & 48.51                       & 72.16                   & 66.00                      & 73.97                         & 25.64                          & 50.94                     & 54.96                         & 58.71                      & 70.20                   & 65.65                          & 56.04                          & 61.69                    & 69.80                   & 30.33                     & 61.54                        & 59.22                   \\
LendingClub-3          & 43.17                          & 93.98                           & 45.53                       & 74.54                   & 61.50                      & 74.39                         & 21.96                          & 50.65                     & 55.16                         & 58.36                      & 76.73                   & 65.07                          & 53.97                          & 61.29                    & 73.80                   & 30.00                     & 61.36                        & 58.91                   \\
LendingClub-4          & 43.17                          & 97.40                           & 49.36                       & 73.51                   & 35.50                      & 71.99                         & 28.88                          & 51.08                     & 55.23                         & 58.27                      & 69.17                   & 65.80                          & 55.52                          & 61.90                    & 69.80                   & 30.00                     & 60.26                        & 57.46                   \\
LendingClub-5          & 43.17                          & 89.15                           & 45.11                       & 75.05                   & 56.00                      & 73.01                         & 24.59                          & 50.86                     & 55.10                         & 58.71                      & 70.90                   & 64.93                          & 55.42                          & 62.10                    & 70.40                   & 29.67                     & 60.07                        & 57.90                   \\
LendingClub-6          & 43.17                          & 93.57                           & 51.91                       & 73.40                   & 40.00                      & 73.84                         & 25.09                          & 50.48                     & 55.23                         & 58.53                      & 69.51                   & 64.64                          & 55.11                          & 62.50                    & 72.20                   & 30.00                     & 60.26                        & 57.61                   \\
FiQA-SA-1              & 43.17                          & 80.82                           & 78.30                       & 36.91                   & 63.50                      & 59.95                         & 26.46                          & 50.83                     & 55.03                         & 52.41                      & 74.95                   & 56.81                          & 51.60                          & 38.71                    & 57.00                   & 15.33                     & 32.23                        & 51.41                   \\
FiQA-SA-2              & 43.17                          & 80.79                           & 71.49                       & 40.52                   & 66.50                      & 59.95                         & 19.39                          & 51.13                     & 55.30                         & 54.86                      & 70.83                   & 53.04                          & 50.77                          & 41.33                    & 58.60                   & 15.67                     & 35.53                        & 51.11                   \\
FiQA-SA-3              & 43.17                          & 80.82                           & 78.72                       & 29.38                   & 66.00                      & 59.95                         & 8.26                           & 51.10                     & 53.53                         & 53.89                      & 67.57                   & 56.38                          & 49.23                          & 40.93                    & 59.60                   & 16.33                     & 34.98                        & 49.99                   \\
FiQA-SA-4              & 43.17                          & 80.82                           & 85.11                       & 28.66                   & 64.50                      & 59.95                         & 3.00                           & 51.40                     & 53.74                         & 55.64                      & 71.15                   & 66.67                          & 51.91                          & 41.73                    & 60.40                   & 18.67                     & 33.52                        & 51.18                   \\
FiQA-SA-5              & 43.17                          & 80.82                           & 71.91                       & 40.10                   & 63.00                      & 59.95                         & 14.61                          & 51.67                     & 54.14                         & 50.31                      & 77.13                   & 56.23                          & 48.92                          & 35.89                    & 56.80                   & 17.33                     & 30.77                        & 50.16                   \\
FiQA-SA-6              & 43.17                          & 80.82                           & 69.36                       & 36.49                   & 65.50                      & 59.95                         & 5.68                           & 51.24                     & 53.67                         & 52.32                      & 76.18                   & 56.38                          & 50.36                          & 35.48                    & 57.00                   & 15.67                     & 32.23                        & 49.50                   \\
FiQA-SA-7              & 43.17                          & 80.82                           & 78.72                       & 29.38                   & 62.00                      & 59.95                         & 9.14                           & 51.32                     & 52.65                         & 54.68                      & 71.70                   & 55.65                          & 46.34                          & 37.50                    & 57.00                   & 20.33                     & 32.23                        & 49.56                   \\
FiQA-SA-8              & 43.17                          & 80.82                           & 77.45                       & 30.72                   & 64.50                      & 59.95                         & 8.55                           & 52.02                     & 54.14                         & 56.87                      & 74.72                   & 59.28                          & 52.43                          & 35.28                    & 59.00                   & 17.67                     & 29.30                        & 50.35                   \\
FPB-1                  & 43.17                          & 80.82                           & 50.64                       & 82.58                   & 36.50                      & 77.51                         & 24.66                          & 50.56                     & 55.84                         & 58.27                      & 83.63                   & 65.07                          & 58.10                          & 62.90                    & 77.60                   & 35.33                     & 61.90                        & 59.12                   \\
FPB-2                  & 43.17                          & 80.82                           & 52.34                       & 84.43                   & 40.00                      & 78.25                         & 23.50                          & 50.43                     & 56.05                         & 57.74                      & 83.98                   & 63.91                          & 56.76                          & 63.31                    & 78.00                   & 35.33                     & 64.29                        & 59.55                   \\
FPB-3                  & 43.17                          & 80.82                           & 50.21                       & 85.57                   & 35.50                      & 78.10                         & 21.93                          & 50.38                     & 55.71                         & 58.18                      & 84.21                   & 63.62                          & 58.62                          & 62.30                    & 77.20                   & 35.67                     & 64.10                        & 59.13                   \\
FPB-4                  & 43.17                          & 80.82                           & 48.94                       & 84.43                   & 35.50                      & 77.94                         & 22.62                          & 50.51                     & 55.91                         & 57.48                      & 84.86                   & 64.64                          & 58.82                          & 62.30                    & 77.60                   & 36.00                     & 63.55                        & 59.12                   \\
German-1               & 43.17                          & 80.82                           & 37.45                       & 74.02                   & 66.00                      & 72.25                         & 27.86                          & 50.81                     & 55.43                         & 58.36                      & 71.12                   & 64.64                          & 51.50                          & 60.48                    & 73.20                   & 34.67                     & 60.62                        & 57.79                   \\
German-2               & 43.17                          & 80.82                           & 38.72                       & 73.81                   & 69.50                      & 71.80                         & 24.71                          & 51.13                     & 55.37                         & 58.18                      & 71.02                   & 65.07                          & 51.91                          & 59.48                    & 74.80                   & 35.00                     & 61.72                        & 58.01                   \\
German-3               & 43.17                          & 80.82                           & 34.04                       & 75.46                   & 67.50                      & 74.55                         & 23.79                          & 50.94                     & 55.37                         & 58.62                      & 69.92                   & 65.51                          & 51.19                          & 60.28                    & 76.20                   & 35.00                     & 61.72                        & 57.89                   \\
German-4               & 43.17                          & 80.82                           & 34.47                       & 76.19                   & 67.00                      & 74.15                         & 26.38                          & 51.16                     & 55.37                         & 58.36                      & 72.47                   & 64.78                          & 54.08                          & 60.08                    & 74.00                   & 35.00                     & 60.62                        & 58.12                   \\
German-5               & 43.17                          & 80.82                           & 33.19                       & 75.15                   & 66.00                      & 72.10                         & 28.97                          & 51.02                     & 55.03                         & 58.62                      & 76.96                   & 65.07                          & 54.39                          & 60.48                    & 67.40                   & 33.33                     & 60.99                        & 57.81                   \\
German-6               & 43.17                          & 80.82                           & 35.74                       & 74.12                   & 66.00                      & 69.66                         & 26.64                          & 50.94                     & 55.10                         & 57.92                      & 78.55                   & 65.51                          & 54.18                          & 61.09                    & 66.00                   & 32.00                     & 61.17                        & 57.57                   \\
German-7               & 43.17                          & 80.82                           & 35.74                       & 76.19                   & 68.00                      & 74.32                         & 21.69                          & 51.34                     & 55.37                         & 58.79                      & 75.57                   & 66.09                          & 54.70                          & 60.48                    & 73.40                   & 34.33                     & 60.81                        & 58.28                   \\
German-8               & 43.17                          & 80.82                           & 35.32                       & 75.77                   & 66.50                      & 74.08                         & 25.28                          & 51.10                     & 55.64                         & 58.79                      & 75.09                   & 65.65                          & 54.28                          & 60.69                    & 72.00                   & 34.33                     & 60.26                        & 58.16                   \\
Headlines-1            & 42.45                          & 80.82                           & 33.62                       & 78.56                   & 65.50                      & 93.21                         & 23.77                          & 50.67                     & 54.96                         & 58.01                      & 85.16                   & 62.03                          & 49.23                          & 61.90                    & 76.80                   & 38.67                     & 60.26                        & 59.74                   \\
Headlines-2            & 41.73                          & 80.82                           & 32.77                       & 78.04                   & 65.00                      & 95.87                         & 24.80                          & 50.70                     & 55.84                         & 58.01                      & 89.74                   & 56.81                          & 48.40                          & 62.10                    & 80.60                   & 36.00                     & 57.88                        & 59.71                   \\
Headlines-3            & 42.45                          & 80.82                           & 29.79                       & 76.60                   & 65.50                      & 97.76                         & 23.36                          & 50.75                     & 55.77                         & 57.92                      & 90.94                   & 54.78                          & 48.40                          & 60.08                    & 80.40                   & 40.33                     & 57.69                        & 59.61                   \\
NER-1                  & 43.17                          & 80.82                           & 42.98                       & 75.57                   & 36.50                      & 68.58                         & 57.82                          & 50.65                     & 55.50                         & 58.09                      & 67.66                   & 64.78                          & 55.11                          & 60.28                    & 72.20                   & 29.67                     & 61.72                        & 57.71                   \\
NER-2                  & 43.17                          & 80.82                           & 41.28                       & 75.36                   & 46.00                      & 72.91                         & 47.07                          & 51.21                     & 55.37                         & 58.79                      & 68.88                   & 64.06                          & 50.46                          & 61.09                    & 75.60                   & 37.00                     & 60.81                        & 58.23                   \\
NER-3                  & 43.17                          & 80.38                           & 52.34                       & 71.96                   & 33.50                      & 64.07                         & 59.42                          & 50.75                     & 55.30                         & 57.39                      & 56.43                   & 63.77                          & 52.43                          & 61.49                    & 70.40                   & 30.00                     & 59.71                        & 56.62                   \\
NER-4                  & 43.17                          & 80.82                           & 39.15                       & 75.57                   & 56.50                      & 73.58                         & 46.84                          & 51.29                     & 55.57                         & 58.62                      & 69.82                   & 64.93                          & 49.23                          & 60.89                    & 75.40                   & 37.00                     & 60.81                        & 58.78                   \\
ACL18-1                & 43.17                          & 80.71                           & 37.45                       & 75.98                   & 64.00                      & 74.08                         & 26.27                          & 52.69                     & 44.63                         & 43.57                      & 70.39                   & 67.39                          & 47.88                          & 60.08                    & 80.00                   & 35.67                     & 60.81                        & 56.75                   \\
ACL18-2                & 43.17                          & 79.67                           & 44.68                       & 76.80                   & 64.50                      & 71.52                         & 20.12                          & 52.77                     & 45.65                         & 46.81                      & 67.14                   & 67.25                          & 49.43                          & 60.69                    & 77.80                   & 32.67                     & 62.09                        & 56.63                   \\
ACL18-3                & 43.17                          & 80.82                           & 40.00                       & 75.88                   & 48.00                      & 72.68                         & 27.12                          & 53.82                     & 47.08                         & 50.22                      & 68.18                   & 68.12                          & 48.50                          & 61.29                    & 77.00                   & 33.33                     & 61.17                        & 56.26                   \\
ACL18-4                & 43.17                          & 79.67                           & 47.23                       & 72.16                   & 66.00                      & 71.25                         & 26.94                          & 50.65                     & 44.16                         & 41.82                      & 66.91                   & 67.54                          & 47.68                          & 60.08                    & 78.40                   & 35.00                     & 59.34                        & 56.35                   \\
ACL18-5                & 43.17                          & 79.26                           & 43.83                       & 76.19                   & 40.00                      & 73.78                         & 26.15                          & 55.99                     & 47.15                         & 53.81                      & 70.18                   & 65.80                          & 48.81                          & 61.09                    & 79.40                   & 34.33                     & 61.17                        & 56.48                   \\
ACL18-6                & 43.17                          & 79.38                           & 41.70                       & 73.30                   & 53.50                      & 70.80                         & 26.99                          & 52.10                     & 46.94                         & 53.98                      & 70.41                   & 67.54                          & 47.57                          & 60.89                    & 80.40                   & 31.67                     & 59.89                        & 56.48                   \\
ACL18-7                & 43.17                          & 80.82                           & 37.45                       & 76.39                   & 35.50                      & 74.78                         & 28.80                          & 52.39                     & 50.14                         & 51.62                      & 73.41                   & 67.10                          & 48.81                          & 60.28                    & 78.80                   & 34.33                     & 61.54                        & 56.20                   \\
ACL18-8                & 43.17                          & 80.82                           & 31.91                       & 75.67                   & 48.50                      & 75.08                         & 25.14                          & 52.12                     & 50.75                         & 53.54                      & 77.20                   & 67.10                          & 49.23                          & 61.49                    & 81.80                   & 32.33                     & 62.45                        & 56.96                   \\
BigData22-1            & 43.17                          & 80.82                           & 44.26                       & 74.54                   & 36.50                      & 71.57                         & 23.29                          & 52.07                     & 49.18                         & 56.17                      & 67.37                   & 68.26                          & 52.22                          & 60.69                    & 77.60                   & 33.00                     & 61.54                        & 56.01                   \\
BigData22-2            & 43.17                          & 80.82                           & 58.30                       & 72.37                   & 35.50                      & 68.77                         & 27.98                          & 52.15                     & 57.54                         & 52.93                      & 59.23                   & 68.84                          & 51.91                          & 58.47                    & 71.00                   & 33.33                     & 60.26                        & 56.03                   \\
BigData22-3            & 43.17                          & 78.86                           & 66.38                       & 70.52                   & 39.00                      & 65.30                         & 26.94                          & 57.42                     & 51.09                         & 51.88                      & 60.30                   & 68.55                          & 50.05                          & 56.45                    & 78.60                   & 30.67                     & 58.06                        & 56.07                   \\
BigData22-4            & 43.17                          & 80.82                           & 42.13                       & 74.95                   & 37.50                      & 71.07                         & 21.00                          & 52.50                     & 49.59                         & 54.16                      & 66.49                   & 68.12                          & 54.70                          & 61.29                    & 75.60                   & 33.67                     & 61.36                        & 55.77                   \\
BigData22-5            & 43.17                          & 80.82                           & 56.60                       & 73.92                   & 37.00                      & 69.57                         & 26.68                          & 52.23                     & 50.75                         & 52.49                      & 61.07                   & 67.39                          & 52.43                          & 60.48                    & 76.00                   & 32.33                     & 61.17                        & 56.12                   \\
BigData22-6            & 43.17                          & 80.01                           & 63.83                       & 72.06                   & 64.00                      & 65.69                         & 25.79                          & 58.06                     & 54.28                         & 52.58                      & 53.43                   & 69.57                          & 50.98                          & 60.69                    & 74.00                   & 34.33                     & 61.72                        & 57.89                   \\
BigData22-7            & 43.17                          & 80.82                           & 41.70                       & 75.36                   & 35.00                      & 70.54                         & 24.25                          & 52.42                     & 51.15                         & 53.72                      & 63.77                   & 68.41                          & 51.70                          & 62.10                    & 72.40                   & 34.33                     & 61.17                        & 55.41                   \\
BigData22-8            & 43.17                          & 80.82                           & 46.81                       & 74.85                   & 37.50                      & 71.41                         & 26.87                          & 51.96                     & 53.74                         & 56.26                      & 68.41                   & 66.96                          & 49.95                          & 60.89                    & 75.60                   & 34.33                     & 60.62                        & 56.48                   \\
BigData22-9            & 43.88                          & 79.38                           & 66.38                       & 67.94                   & 54.00                      & 67.05                         & 24.39                          & 53.84                     & 54.28                         & 51.97                      & 59.03                   & 69.13                          & 47.99                          & 60.48                    & 71.40                   & 32.00                     & 60.26                        & 56.67                   \\
CIKM18-1               & 43.17                          & 80.82                           & 38.30                       & 75.67                   & 41.50                      & 73.95                         & 21.87                          & 50.54                     & 53.94                         & 57.13                      & 73.64                   & 65.07                          & 53.25                          & 60.48                    & 79.40                   & 33.33                     & 60.26                        & 56.61                   \\
CIKM18-2               & 43.17                          & 80.82                           & 40.85                       & 74.74                   & 39.50                      & 73.26                         & 24.57                          & 50.89                     & 52.11                         & 56.87                      & 72.30                   & 66.96                          & 53.04                          & 59.68                    & 80.60                   & 33.67                     & 60.62                        & 56.69                   \\
CIKM18-3               & 43.17                          & 80.82                           & 43.83                       & 74.74                   & 37.00                      & 73.16                         & 24.15                          & 51.80                     & 51.29                         & 56.43                      & 69.89                   & 65.22                          & 52.84                          & 60.69                    & 79.40                   & 35.00                     & 60.44                        & 56.46                   \\
CIKM18-4               & 43.17                          & 80.82                           & 37.87                       & 75.57                   & 38.50                      & 73.85                         & 22.13                          & 50.94                     & 54.01                         & 58.09                      & 70.19                   & 64.93                          & 52.43                          & 61.29                    & 79.20                   & 35.00                     & 60.44                        & 56.38                   \\
CIKM18-5               & 43.17                          & 80.82                           & 42.55                       & 74.54                   & 44.00                      & 72.90                         & 23.28                          & 50.27                     & 52.51                         & 56.87                      & 70.86                   & 66.96                          & 52.12                          & 59.27                    & 78.60                   & 33.33                     & 60.44                        & 56.62                   \\
CIKM18-6               & 43.17                          & 80.82                           & 47.23                       & 73.92                   & 35.00                      & 72.31                         & 25.37                          & 51.16                     & 52.45                         & 56.43                      & 69.15                   & 65.65                          & 52.32                          & 59.88                    & 79.20                   & 34.67                     & 60.44                        & 56.42                   \\
CIKM18-7               & 43.17                          & 80.82                           & 40.85                       & 75.46                   & 38.00                      & 73.25                         & 24.52                          & 50.43                     & 54.08                         & 56.34                      & 71.99                   & 65.22                          & 54.28                          & 60.48                    & 78.00                   & 34.33                     & 59.71                        & 56.53                   \\
CIKM18-8               & 43.17                          & 80.82                           & 43.83                       & 74.33                   & 39.50                      & 73.66                         & 21.59                          & 50.40                     & 51.56                         & 56.43                      & 68.15                   & 65.65                          & 52.43                          & 60.89                    & 78.40                   & 35.00                     & 60.62                        & 56.26                   \\
CIKM18-9               & 43.17                          & 80.82                           & 48.94                       & 74.33                   & 39.00                      & 72.72                         & 25.82                          & 51.08                     & 51.43                         & 54.16                      & 66.47                   & 65.22                          & 50.88                          & 59.88                    & 78.60                   & 35.67                     & 61.54                        & 56.45                   \\
CIKM18-10              & 43.17                          & 80.82                           & 49.36                       & 71.55                   & 41.00                      & 72.09                         & 23.48                          & 50.81                     & 51.97                         & 57.22                      & 76.32                   & 66.96                          & 52.94                          & 58.87                    & 81.60                   & 32.33                     & 59.71                        & 57.07                   \\
CIKM18-11              & 43.17                          & 77.11                           & 55.74                       & 71.86                   & 41.00                      & 72.02                         & 26.59                          & 50.56                     & 50.07                         & 56.52                      & 78.01                   & 68.26                          & 51.81                          & 58.06                    & 84.00                   & 33.33                     & 58.24                        & 57.43                   \\
CIKM18-12              & 43.17                          & 80.82                           & 34.47                       & 75.77                   & 35.50                      & 74.66                         & 19.54                          & 50.08                     & 53.80                         & 56.26                      & 74.01                   & 65.36                          & 48.71                          & 60.69                    & 80.60                   & 33.33                     & 60.81                        & 55.74                   \\
CIKM18-13              & 43.17                          & 80.82                           & 33.62                       & 75.77                   & 40.50                      & 74.54                         & 22.61                          & 50.73                     & 53.60                         & 56.52                      & 71.36                   & 65.65                          & 49.23                          & 60.48                    & 79.80                   & 34.67                     & 61.54                        & 56.15                   \\
CIKM18-14              & 43.17                          & 76.25                           & 54.89                       & 70.52                   & 40.00                      & 68.28                         & 30.18                          & 51.53                     & 51.97                         & 53.37                      & 70.77                   & 66.09                          & 50.57                          & 60.89                    & 80.20                   & 33.33                     & 58.97                        & 56.53                   \\
CIKM18-15              & 43.17                          & 80.82                           & 34.04                       & 75.98                   & 38.00                      & 74.33                         & 21.76                          & 50.70                     & 54.08                         & 56.96                      & 67.90                   & 65.22                          & 48.30                          & 59.88                    & 79.20                   & 35.33                     & 61.90                        & 55.74                   \\
CIKM18-16              & 43.17                          & 80.79                           & 44.68                       & 72.78                   & 41.50                      & 71.64                         & 25.79                          & 52.34                     & 52.99                         & 49.96                      & 68.00                   & 66.67                          & 50.15                          & 61.49                    & 80.20                   & 33.33                     & 58.79                        & 56.13                   \\
CIKM18-17              & 43.17                          & 80.82                           & 33.62                       & 75.88                   & 40.50                      & 73.96                         & 21.15                          & 50.32                     & 54.48                         & 56.34                      & 73.22                   & 65.80                          & 49.85                          & 60.48                    & 79.80                   & 34.33                     & 60.44                        & 56.13                   \\
\bottomrule
\end{tabular}
}
\end{table}

\begin{table}[t]
\centering
\renewcommand{\arraystretch}{1.2}
\caption{Performance values of the three base models on the 34 tasks of FinBen. The name of the datasets we select for evaluation is emphasized in bold.}
\label{tab:34tasks}
\begin{tabular}{ll|ccc}
\toprule
\multicolumn{2}{c}{User Task} & \multicolumn{3}{c}{Base Models} \\
\cmidrule(lr){1-2}\cmidrule(lr){3-5}
Dataset              & Metric    & Qwen2.5-7B & Llama3.1-8B-Instruct & Llama3.1-8B \\
\midrule
\textbf{Australian}  & Acc       & 43.17      & 44.60                & 43.17       \\
\textbf{SC}          & Acc       & 65.14      & 88.48                & 79.45       \\
CD                   & EntityF1  & 0.00       & 0.00                 & 0.00        \\
ConvFinQA            & Acc       & 0.13       & 0.00                 & 0.00        \\
ccf                  & Acc       & 99.87      & 3.29                 & 0.57        \\
ccfraud              & Acc       & 92.33      & 55.39                & 93.80       \\
\textbf{LendingClub} & Acc       & 80.82      & 76.33                & 57.34       \\
polish               & Acc       & 94.53      & 77.78                & 94.88       \\
portoseguro          & Acc       & 96.98      & 18.47                & 96.47       \\
taiwan               & Acc       & 96.70      & 95.53                & 95.02       \\
travelinsurance      & Acc       & 98.50      & 1.50                 & 1.50        \\
ECTSUM               & BartScore & -517.76    & -517.76              & -517.76     \\
EDTSUM               & BartScore & -381.09    & -378.12              & -429.06     \\
\textbf{FinArg-ARC}  & Acc       & 64.78      & 46.67                & 60.00       \\
\textbf{FinArg-ACC}  & Acc       & 48.30      & 51.81                & 49.85       \\
FINER-ORD            & EntityF1  & 0.35       & 3.88                 & 0.17        \\
FinQA                & Acc       & 0.09       & 0.00                 & 0.00        \\
FinRED               & F1        & 0.00       & 0.01                 & 0.02        \\
\textbf{FiQA-SA}     & Acc       & 38.30      & 40.43                & 56.17       \\
FNXL                 & EntityF1  & 0.00       & 0.00                 & 0.00        \\
\textbf{FOMC}        & Acc       & 60.48      & 29.44                & 34.68       \\
\textbf{FPB}         & Acc       & 76.08      & 32.78                & 30.72       \\
FSRL                 & EntityF1  & 0.23       & 0.00                 & 0.00        \\
\textbf{German}      & Acc       & 65.00      & 49.50                & 66.00       \\
\textbf{Headlines}   & AvgF1     & 74.81      & 59.95                & 59.95       \\
\textbf{MA}          & Acc       & 79.20      & 56.40                & 51.00       \\
\textbf{MLESG}       & Acc       & 35.67      & 32.67                & 20.00       \\
\textbf{MultiFin}    & Acc       & 60.99      & 31.32                & 28.39       \\
\textbf{NER}         & EntityF1  & 21.75      & 0.62                 & 9.01        \\
\textbf{ACL18}       & Acc       & 51.10      & 51.40                & 51.34       \\
\textbf{BigData22}   & Acc       & 55.30      & 55.57                & 52.79       \\
\textbf{CIKM18}      & Acc       & 58.44      & 54.24                & 54.07       \\
TATQA                & Acc       & 12.95      & 0.00                 & 0.00        \\
TSA                  & RMSE (↓)  & 36.82      & 15.93                & 30.92   \\
\bottomrule
\end{tabular}
\end{table}

\paragraph{Evaluation Benchmark.} We select several datasets of the FinBen to serve as our financial evaluation benchmark. It contains the 10 datasets we have mentioned above and other 7 datasets: Information Extraction(IE) task, SC; Textual Analysis(TA) tasks, FOMC, FinArg-ACC, FinArg-ARC, MultiFin, MA, MLESG. Since the version of \texttt{lm-evaluation-harness} used by FinBen's official task evaluation implementation code \footnote{\url{https://github.com/The-FinAI/PIXIU}} is outdated, in order to improve the efficiency of the evaluation, we re-implemented the tasks we needed using the new version of \texttt{lm-evaluation-harness}, which do not show much difference in performance compared to the origin code. The criteria for selecting 17 tasks from the original 34 tasks are as follows: We first evaluate the three base model on all the 34 tasks in FinBen and select the task with at least one base model has a score between 20-90 (out of 100). For task with more than one metric, we select one metric value as the score, which is either Acc, F1, RMSE or BART score. All results presented are in the zero-shot setting. See \Cref{tab:34tasks} for a complete performance table of the three base models on the 34 tasks.

\subsubsection{Healthcare}
\paragraph{Model hub details.} For the instruction tuning datasets used to train models, we use the training data of some famous open-sourced medical LLMs. Specifically, the tuning datasets are listed as follows: (1) MedInstruct-52K \citep{zhang2023alpacare}: A multi-task dataset for the medical field, integrating multiple sources from the medical field, including medical literature, clinical records, and guidelines, making it ideal for training models to generate and understand medical tasks and questions. This dataset is originally used to train the Alpacare \citep{zhang2023alpacare} model. The dataset path in Hugging Face is \textit{lavita/AlpaCare-MedInstruct-52k}. (2) ChatDoctor-100K \citep{li2023chatdoctor}: A medical conversation dataset, collected based on real patient-doctor interactions, including a wealth of terms, knowledge and expertise essential for training LLMs in the medical domain. The dataset path in Hugging Face is \textit{lavita/ChatDoctor-HealthCareMagic-100k}. (3) Medical Meadow \citep{han2023medalpaca}: A collection of medical tasks for fine-tuning, containing multiple parts. We use the parts available in Hugging Face, including Wikidoc, Medical Flashcards, Wikidoc Patient Information, PubMed Causal, MEDIQA, Health Advice and MedQA. To avoid test data leakage, we exclude the medical part of MMLU. We also exclude the CORD-19 dataset, which is too large and seemly little relevant to the evaluation tasks. It is worth noting that Medical Meadow includes the training data from the MedQA benchmark, whose test split is used to do evaluation. Therefore, to fine-tune more models, we also use the training split of another evaluation task, MedMCQA. We also use the combination of some of the above datasets to do fine-tuning. In summary, all combinations we use is listed in \Cref{tab:med-name}.

\begin{table}[t]
\centering
\small
\renewcommand{\arraystretch}{1.2}
\caption{All the fine-tuning dataset name we use and its source in healthcare scenario.}
\label{tab:med-name}
\begin{tabular}{@{}c|c@{}}
\toprule
Fine-tuning task name                    & Source                                                                           \\ \midrule
AlpaCare                                 & MedInstruct-52K                                                                  \\
ChatDoctor                               & ChatDoctor-100K                                                                  \\
medalpaca\_cleaned                       & Medical Meadow (excluding MMLU and CORD-19)                                      \\
medical\_flashcards                      & Medical Flashcards                                                               \\
pubmed\_causal                           & PubMed Causal                                                                    \\
medqa\_train                             & Training split of MedQA                                                          \\
medmcqa\_train                           & Training split of MedMCQA                                                        \\
AlpaCare\&ChatDoctor                     & MedInstruct-52K and ChatDoctor-100K                                              \\
medalpaca\_cleaned\&AlpaCare\&ChatDoctor & MedInstruct-52K, ChatDoctor-100K and                                             \\
                                         & Medical Meadow (excluding MMLU and CORD-19)                                      \\
medqa\_train\&pubmed\_causal             & MedQA-train and PubMed Causal                                                    \\
medqa\_train\&medmcqa\_train             & MedQA-train and MedMCQA-train                                                    \\ \bottomrule
\end{tabular}
\end{table}

Similarly, we first evaluate the three base models and get their scores. Qwen2.5-7B almost achieve the best score on all datasets except \texttt{clinical\_knowledge}. Therefore, we only use Qwen2.5-7B to be the base model of fine-tuning. Finally, we construct a model hub with 13 models. See \Cref{tab:med-models} for a complete performance table of the three base models and all the models in model hub.

\begin{table}[t]
\centering
\renewcommand{\arraystretch}{1.5}
\caption{Performance value of all models in model hub in healthcare scenario.}
\label{tab:med-models}
\resizebox{\textwidth}{!}{
\begin{tabular}{c|ccccccccc|c}
\toprule
\diagbox{Models}{Tasks}                                   & medmcqa & medqa\_4options & anatomy & clinical\_knowledge & college\_biology & college\_medicine & medical\_genetics & professional\_medicine & pubmedqa & \textbf{Avg} \\
\midrule
Llama-3.1-8B                               & 56.54   & 59.86           & 65.19   & 72.83               & 77.78            & 63.58             & 84.00             & 73.06                  & 75.80    & 69.85        \\
Llama-3.1-8B-Instruct                      & 59.26   & 63.63           & 68.15   & 78.87               & 81.25            & 68.79             & 77.00             & 77.57                  & 74.80    & 72.15        \\
Qwen2.5-7B                                 & 59.93   & 64.18           & 71.85   & 77.36               & 82.64            & 69.36             & 87.00             & 78.68                  & 75.20    & 74.02        \\
\midrule
medqa\_train\&pubmed\_causal-1             & 59.48   & 65.59           & 71.85   & 77.36               & 86.11            & 68.79             & 87.00             & 76.84                  & 76.00    & 74.34        \\
medqa\_train-1                             & 59.48   & 65.59           & 71.85   & 77.74               & 84.72            & 69.94             & 88.00             & 79.78                  & 75.80    & 74.77        \\
pubmed\_causal-1                           & 60.32   & 63.55           & 71.85   & 78.11               & 81.94            & 69.36             & 88.00             & 77.57                  & 76.80    & 74.17        \\
medalpaca\_cleaned-1                       & 59.81   & 63.86           & 70.37   & 78.87               & 84.72            & 69.94             & 85.00             & 77.94                  & 75.80    & 74.03        \\
medqa\_train\&medmcqa\_train-1    & 62.49   & 64.81           & 70.37   & 78.49               & 84.03            & 68.79             & 89.00             & 78.68                  & 76.80    & 74.83        \\
medmcqa\_train-1                           & 62.01   & 63.63           & 71.11   & 79.25               & 85.42            & 68.21             & 89.00             & 76.47                  & 75.80    & 74.54        \\
AlpaCare-1                                 & 59.77   & 62.92           & 71.85   & 78.49               & 84.03            & 68.79             & 87.00             & 77.21                  & 75.00    & 73.90        \\
ChatDoctor-1 & 60.29  & 63.63  & 72.59  & 77.74  & 84.03  & 67.05  & 86.00  & 77.57  & 74.80  & 73.74  \\ 
ChatDoctor-2 & 60.15  & 63.32  & 73.33  & 76.60  & 81.94  & 67.63  & 88.00  & 77.21  & 73.80  & 73.55  \\ 
AlpaCare\&ChatDoctor-1 & 58.93  & 62.14  & 70.37  & 78.49  & 82.64  & 69.36  & 86.00  & 75.74  & 74.80  & 73.16  \\ 
AlpaCare\&ChatDoctor-2 & 58.38  & 61.67  & 70.37  & 78.11  & 82.64  & 68.79  & 87.00  & 76.10  & 75.00  & 73.12  \\ 
medalpaca\_cleaned\&AlpaCare\&ChatDoctor-1 & 59.72  & 62.61  & 70.37  & 78.11  & 84.72  & 68.79  & 85.00  & 77.21  & 76.20  & 73.64  \\ 
medalpaca\_cleaned\&AlpaCare\&ChatDoctor-2 & 59.55  & 62.37  & 71.11  & 77.74  & 86.11  & 71.10  & 83.00  & 76.84  & 75.60  & 73.71 \\ 
\bottomrule
\end{tabular}
}
\end{table}

\paragraph{Evaluation Benchmark.} To ensure the authority, we use the Open Medical LLM Leaderboard \citep{OpenMedicalLLMLeaderboard,singhal2022large} to be the evaluation benchmark in medical scenario. The leaderboard, widely recognized within the Hugging Face community, is designed to monitor, rank, and assess the efficacy of large language models (LLMs) in addressing medical question-answering challenges. It evaluates LLMs across 9 medical datasets, including MedQA (USMLE), PubMedQA, MedMCQA, and 6 subsets of MMLU related to medicine and biology. The leaderboard provides a detailed evaluation of each model’s proficiency in medical knowledge and question-answering abilities. All results presented are in the zero-shot setting and accuracy is employed as the metric. 

\subsubsection{Mathematics}
\paragraph{Model hub details.} For training models, we employ several well-known open-source instruction tuning datasets. The datasets employed for tuning are outlined as follows: (1) MathInstruct \citep{yue2023mammoth}: A math instruction-tuning dataset compiled from 13 math datasets with intermediate rationales, which offers a unique hybrid of chain-of-thought (CoT) and program-of-thought (PoT) rationales and ensures broad coverage of different math fields and complexity levels. (2) MetaMathQA \citep{yu2023metamath}: A diverse and high-quality mathematical dataset augmented from the training sets of GSM8K and MATH. (3) MetaMath-GSM240K: The subset of MetaMath containing the 240,000 samples that were augmented from GSM8K.(4) Arithmo-Data \citep{jindal_2023_arithmo}: the combination of MetaMathQA, MathInstruct and LILA-OOD \citep{mishra2022lila} datasets. (5) Orca-Math-200K \citep{mitra2024orca}: A dataset containing 200,000 grade school math word problems, with answers generated using GPT-4 Turbo. (6) BELLE\_School\_Math\_0.25M \citep{BELLE,sun2023comparative}: A dataset comprising approximately 250,000 Chinese mathematical problems produced by ChatGPT, along with the corresponding problem-solving process. (7) MWP-Instruct: An instruction tuning dataset derived from datasets in MWPToolkit \citep{lan2021mwptoolkit}, containing arithmetic and equation solving tasks. (8) MATH\_train: The training set of MATH \citep{hendrycks2021measuring} dataset. (9) GSM8K\_zh: A dataset for mathematical reasoning in Chinese, question-answer pairs are translated from GSM8K \citep{cobbe2021training} by GPT-3.5-Turbo with few-shot prompting. Only the 7473 training samples are used for supervised fine-tuning.

Through our first evaluation of the three base models, Qwen2.5-7B achieve the best score on all datasets. Thus, we utilize Qwen2.5-7B as the base model for fine-tuning. Ultimately, we establish a model hub comprising 13 models. The performance of all the models is provided in \Cref{tab:math-models}.

\begin{table}[t]
\centering
\setlength{\tabcolsep}{1pt}
\renewcommand{\arraystretch}{1.5}
\caption{Performance value of all models in model hub in mathematics scenario.}
\label{tab:math-models}
\resizebox{\textwidth}{!}{
\begin{tabular}{c|ccccc|ccc}
\toprule
\multirow{2}{*}{\diagbox{Models}{Tasks}}       & \multicolumn{5}{c|}{AGIEval}                                                                                    & \multicolumn{3}{c}{CMMLU}                                                                       \\
                             & agieval\_aqua\_rat & agieval\_gaokao\_mathcloze & agieval\_gaokao\_mathqa & agieval\_math & agieval\_sat\_math & cmmlu\_college\_mathematics & cmmlu\_elementary\_mathematics & cmmlu\_high\_school\_mathematics \\
\midrule
Llama-3.1-8B                 & 21.65              & 3.39                       & 28.77                   & 9.20          & 44.09              & 32.38                       & 31.30                          & 31.71                            \\
Llama-3.1-8B-Instruct        & 27.95              & 2.54                       & 29.06                   & 5.60          & 40.00              & 36.19                       & 29.57                          & 38.41                            \\
Qwen2.5-7B                   & 41.73              & 16.95                      & 49.86                   & 19.80         & 55.91              & 45.71                       & 65.65                          & 61.59                            \\
\midrule
Orca-Math-200K-1             & 40.94              & 12.71                      & 50.71                   & 19.80         & 57.73              & 50.48                       & 65.22                          & 64.63                            \\
Orca-Math-200K-2             & 41.73              & 13.56                      & 51.28                   & 17.20         & 57.27              & 46.67                       & 64.78                          & 64.02                            \\
MWP-Instruct-1               & 41.34              & 11.86                      & 51.57                   & 20.20         & 57.27              & 51.43                       & 64.35                          & 60.98                            \\
GSM8K\_zh-1                  & 38.98              & 7.63                       & 51.57                   & 20.60         & 55.45              & 52.38                       & 66.96                          & 61.59                            \\
MATH\_train-1                & 39.37              & 17.80                      & 50.71                   & 19.60         & 50.00              & 45.71                       & 65.65                          & 62.80                            \\
Arithmo-Data-1               & 38.98              & 13.56                      & 49.86                   & 17.00         & 55.45              & 49.52                       & 65.65                          & 63.41                            \\
MetaMathQA-1                 & 40.55              & 5.93                       & 52.99                   & 28.70         & 55.45              & 48.57                       & 67.39                          & 64.02                            \\
MetaMathQA-2                 & 40.55              & 5.08                       & 53.85                   & 28.10         & 57.27              & 47.62                       & 66.96                          & 62.80                            \\
MetaMath-GSM240K-1           & 41.73              & 14.41                      & 50.14                   & 21.10         & 55.00              & 49.52                       & 64.78                          & 64.63                            \\
BELLE\_School\_Math\_0.25M-1 & 38.19              & 9.32                       & 45.30                   & 16.70         & 54.55              & 50.48                       & 54.35                          & 56.71                            \\
BELLE\_School\_Math\_0.25M-2 & 38.98              & 6.78                       & 45.58                   & 16.90         & 56.82              & 52.38                       & 56.09                          & 57.32                            \\
MathInstruct-1               & 39.37              & 9.32                       & 50.43                   & 18.20         & 55.00              & 47.62                       & 66.09                          & 64.02                            \\
MathInstruct-2               & 40.55              & 17.80                      & 48.43                   & 20.10         & 55.00              & 47.62                       & 65.22                          & 64.63 \\
\bottomrule
\end{tabular}
}

\vspace{12pt}

\resizebox{\textwidth}{!}{
\begin{tabular}{c|c|c|c|c|cccc|c}
\toprule
\multirow{2}{*}{\diagbox{Models}{Tasks}}       & GSM8K  & MathQA & MGSM-zh               & MATH          & \multicolumn{4}{c|}{MMLU}                                                                                               & \multirow{2}{*}{\textbf{Avg}} \\
                             & gsm8k  & mathqa & mgsm\_native\_cot\_zh & minerva\_math & mmlu\_abstract\_algebra & mmlu\_college\_mathematics & mmlu\_elementary\_mathematics & mmlu\_high\_school\_mathematics &                               \\
\midrule
Llama-3.1-8B                 & 50.80 & 39.50  & 9.60                  & 17.92         & 37.00                   & 39.00                      & 43.92                         & 40.74                           & 30.06                         \\
Llama-3.1-8B-Instruct        & 76.12 & 39.30  & 38.40                 & 34.44         & 36.00                   & 34.00                      & 48.68                         & 42.22                           & 34.91                         \\
Qwen2.5-7B                   & 84.08 & 43.32  & 66.40                 & 40.16         & 54.00                   & 53.00                      & 72.75                         & 55.93                           & 51.68                         \\
\midrule
Orca-Math-200K-1             & 84.00 & 45.63  & 67.60                 & 40.90         & 52.00                   & 52.00                      & 72.75                         & 55.19                           & 52.02                         \\
Orca-Math-200K-2             & 83.85 & 46.93  & 70.00                 & 41.56         & 52.00                   & 55.00                      & 73.28                         & 55.19                           & 52.15                         \\
MWP-Instruct-1               & 82.79 & 42.61  & 66.00                 & 39.42         & 55.00                   & 53.00                      & 75.13                         & 55.56                           & 51.78                         \\
GSM8K\_zh-1                  & 74.37 & 36.65  & 73.60                 & 29.64         & 56.00                   & 51.00                      & 74.07                         & 55.93                           & 50.40                         \\
MATH\_train-1                & 76.19 & 41.64  & 66.40                 & 41.96         & 55.00                   & 58.00                      & 73.54                         & 55.93                           & 51.27                         \\
Arithmo-Data-1               & 80.14 & 41.41  & 68.00                 & 45.12         & 53.00                   & 53.00                      & 75.13                         & 55.93                           & 51.57                         \\
MetaMathQA-1                 & 83.85 & 42.38  & 68.00                 & 32.20         & 56.00                   & 53.00                      & 73.81                         & 57.04                           & 51.87                         \\
MetaMathQA-2                 & 83.47 & 42.38  & 68.80                 & 30.32         & 54.00                   & 55.00                      & 73.02                         & 56.67                           & 51.62                         \\
MetaMath-GSM240K-1           & 84.15 & 41.17  & 71.60                 & 36.48         & 54.00                   & 56.00                      & 73.02                         & 55.56                           & 52.08                         \\
BELLE\_School\_Math\_0.25M-1 & 81.50 & 40.13  & 61.20                 & 43.14         & 51.00                   & 48.00                      & 70.90                         & 50.74                           & 48.26                         \\
BELLE\_School\_Math\_0.25M-2 & 82.18 & 40.57  & 57.20                 & 42.40         & 52.00                   & 51.00                      & 71.96                         & 51.11                           & 48.70                         \\
MathInstruct-1               & 79.15 & 40.94  & 65.60                 & 27.58         & 52.00                   & 53.00                      & 73.54                         & 54.81                           & 49.79                         \\
MathInstruct-2               & 80.97 & 41.27  & 68.00                 & 29.94         & 54.00                   & 56.00                      & 74.34                         & 55.19                           & 51.19                        \\
\bottomrule
\end{tabular}
}
\end{table}

\paragraph{Evaluation Benchmark.} To comprehensively assess math capability of the system and models, we refer to the MathEval benchmark \footnote{\url{https://matheval.ai/}} and the evaluation benchmark of DeepSeekMath \citep{shao2024deepseekmath} and collect 16 datasets to serve as our math evaluation benchmark. English datasets include GSM8K \citep{cobbe2021training} (5-shot), MATH \citep{hendrycks2021measuring} (We use the 4-shot version \texttt{minerva\_math}), MathQA \citep{amini2019mathqa}, 4 subsets of MMLU \citep{hendrycks2020measuring} (\texttt{mmlu\_abstract\_algebra}, \texttt{mmlu\_college\_mathematics}, \texttt{mmlu\_elementary\_mathematics}, \texttt{mmlu\_high\_school\_mathematics}) and 3 subsets of AGIEval \citep{zhong2023agieval}: (\texttt{agieval\_math}, \texttt{agieval\_sat\_math}, \texttt{agieval\_aqua\_rat}). Chinese datasets include 2 subsets of AGIEval (\texttt{agieval\_gaokao\_mathcloze}, \texttt{agieval\_gaokao\_mathqa}), 3 subsets of CMMLU \citep{li2023cmmlu}: (\texttt{cmmlu\_college\_mathematics}, \texttt{cmmlu\_high\_school\_mathematics}, \texttt{cmmlu\_elementary\_mathematics}) and Chinese part of MGSM \citep{shi2022language} (We use the CoT version \texttt{mgsm\_native\_cot\_zh}). The evaluation metric employed for GSM8K and MATH tasks is the exact match criterion, while accuracy serves as the evaluation metric for all other tasks.

\subsubsection{Details about LLMs under evaluation}

The base models with parameter under 8B are described as follows:
\begin{itemize}
    \item \textbf{Llama3.1-8B} \citep{llama3}. Developed by Meta AI, Llama3.1-8B is part of the Llama 3.1 series and comprises 8 billion parameters. It supports a context length of up to 128,000 tokens, enabling it to process extensive textual inputs effectively. The model is multilingual, supporting languages such as English, German, French, Italian, Portuguese, Hindi, Spanish, and Thai. It incorporates grouped query attention (GQA) mechanisms, which enhance its ability to handle longer contexts efficiently.
    \item \textbf{Llama3.1-8B-Instruct} \citep{llama3}. This variant of the Llama3.1-8B model has undergone instruction fine-tuning, enhancing its ability to follow user prompts more accurately and generate contextually appropriate responses. 
    \item \textbf{Qwen2.5-7B} \citep{yang2024qwen2.5}. Released by Alibaba, Qwen2.5-7B is a 7-billion-parameter model within the Qwen2.5 series. It exhibits significant improvements in knowledge retention, coding, and mathematical capabilities. The model supports a context length of up to 128,000 tokens and can generate texts exceeding 8,000 tokens. It is multilingual, supporting over 29 languages.
\end{itemize}
The large-scale models with parameter over 70B are described as follows.
\begin{itemize}
    \item \textbf{Llama3.1-70B-Instruct} \citep{llama3}. This model is a 70-billion-parameter variant of the Llama 3.1 series, fine-tuned for instruction-based tasks. It supports a context length of up to 128,000 tokens and is designed to understand and respond to user directives with high accuracy. The model is multilingual and incorporates GQA mechanisms to enhance its performance in handling long contexts.
    \item \textbf{Qwen1.5-110B} \citep{qwen1.5}. Qwen1.5-110B is the first model in the Qwen1.5 series to surpass 100 billion parameters, boasting 110 billion parameters. It achieves performance comparable to Meta's Llama3-70B in base model evaluations and excels in chat evaluations, including MT-Bench and AlpacaEval 2.0. The model employs a Transformer decoder architecture with GQA, supports a context length of 32,000 tokens, and is multilingual.
    \item \textbf{Qwen2.5-72B} \citep{yang2024qwen2.5}. As the largest model in the Qwen2.5 series, Qwen2.5-72B contains 72 billion parameters. It surpasses Meta AI's Llama-3-70B in evaluations and offers free commercial use for applications with fewer than 100 million monthly active users. The model supports a context length of up to 128,000 tokens and is suitable for multilingual environments.
    \item \textbf{Flan-PaLM-540B} \citep{chung2024scaling}. Flan-PaLM-540B is a 540-billion-parameter instruction-finetuned language model developed by Google Research, designed to significantly enhance generalization to unseen tasks. Built upon PaLM-540B, it introduces instruction finetuning with 1,836 diverse tasks, covering natural language understanding, reasoning, multilingual question answering, and mathematical problem-solving. 
\end{itemize}

\begin{table}[t]
\renewcommand{\arraystretch}{1.2}
\caption{Comparison of \texttt{Single-Combined} model and our learnware dock system utilizing specialized SLMs on financial evaluation benchmark. Full score is 100. The better score is emphasized in bold.}
\label{tab:fin-extra}
\centering
\begin{tabular}{ll|cc|c}
\toprule
Dataset     & Metric   & Single-Combined & Learnware & Oracle \\
\midrule
Australian  & Acc      & 45.68           & \textbf{56.83}       & 56.83  \\
LendingClub & Acc      & 84.60           & \textbf{92.07}       & 92.07  \\
FiQA-SA     & Acc      & 74.90           & \textbf{76.38}       & 76.38  \\
FPB         & Acc      & 82.17           & \textbf{84.25}       & 84.25  \\
German      & Acc      & 66.00           & \textbf{67.06}       & 67.06  \\
Headlines   & AvgF1    & 92.80           & \textbf{95.61}       & 95.61  \\
NER         & EntityF1 & 49.29           & \textbf{52.79}       & 52.79  \\
ACL18       & Acc      & 49.73           & \textbf{52.82}       & 53.63  \\
BigData22   & Acc      & 49.76           & \textbf{52.40}       & 55.88  \\
CIKM18      & Acc      & 49.46           & \textbf{55.99}       & 58.52  \\
SC          & Acc      & \textbf{88.86}           & 84.17       & 88.61  \\
FinArg-ARC  & Acc      & 62.90           & \textbf{64.31}       & 68.36  \\
FinArg-ACC  & Acc      & 54.41           & \textbf{58.08}       & 58.08  \\
FOMC        & Acc      & \textbf{64.32}           & 62.70       & 62.70  \\
MA          & Acc      & 75.25           & \textbf{79.81}       & 79.81  \\
MLESG       & Acc      & \textbf{35.83}           & 33.42       & 38.33  \\
MultiFin    & Acc      & 59.80           & \textbf{63.46}       & 63.46  \\ \midrule
Avg.         &          & 63.87           & \textbf{66.60}       & 67.79 \\
\bottomrule
\end{tabular}
\end{table}

\subsection{Further discussion in finance scenario}

\subsubsection{Extra exploration experiment}
\label{sec:fin-extra}

In this subsection, we compare our system which utilizes the model hub with combining all available data to fine-tuning a single model with the same size in finance scenario. The results on our financial evaluation benchmark is shown in \Cref{tab:fin-extra}, where the \texttt{Single-Combined} refers to the model fine-tuned using a combination of 10 datasets in FIT. Recall that the models in our model hub are all fine-tuned with a single dataset of FIT. The base model for the fine-tuning of \texttt{Single-Combined} is Qwen2.5-7B.

\subsubsection{Identification results analysis}

\begin{wrapfigure}{r}[0.5cm]{0pt}
  \centering
  \includegraphics[width=0.45\linewidth]{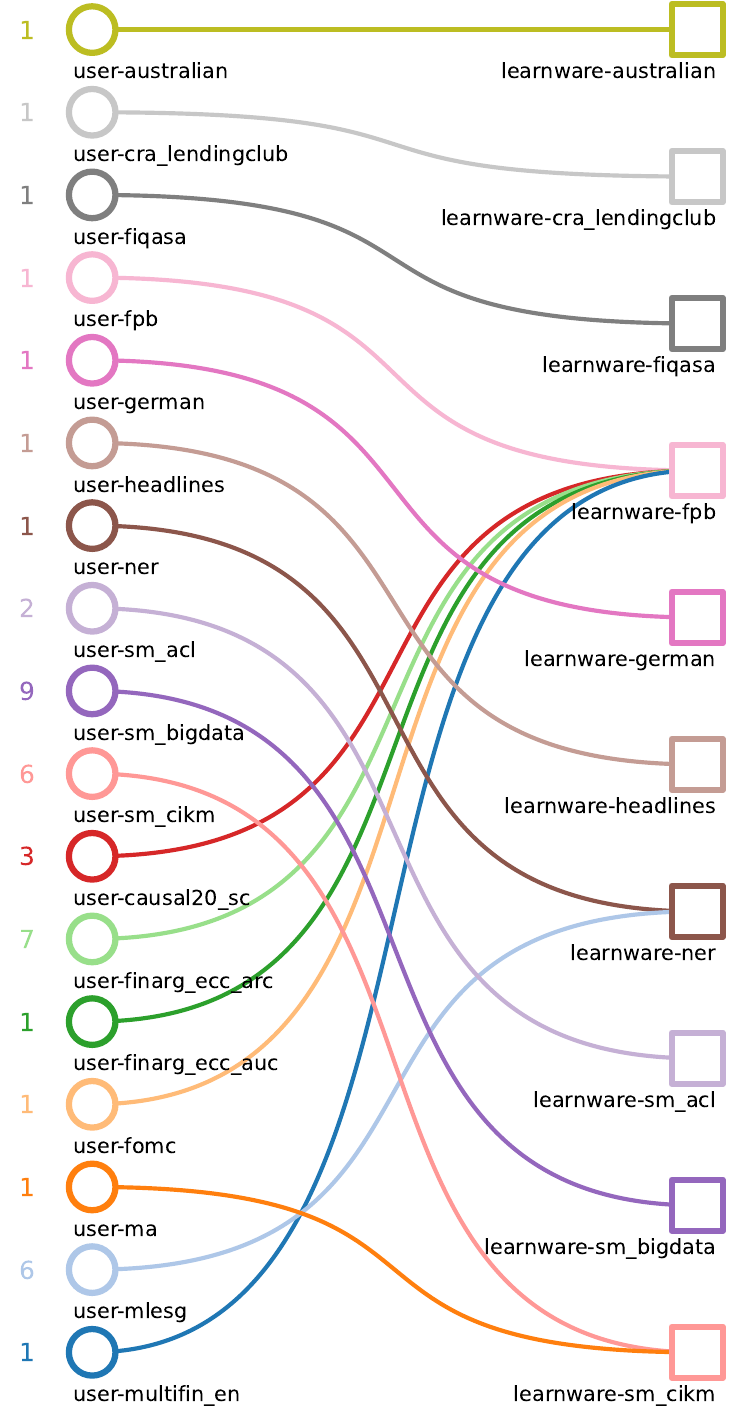}
  \caption{The identification results of our method between user tasks and fine-tuned models. The number on the left of each user task denotes the rank of the selected fine-tuned model's performance among all models in solving this task.}
  \label{fig:finance}
\end{wrapfigure}

\Cref{fig:finance} shows the identification results, linking user tasks with fine-tuned models. The nodes on the left represent user tasks, and the nodes on the right are models trained using the corresponding fine-tuning data. Each line represents our algorithm’s selection results. The number on the left of each user task represents the rank of the selected fine-tuned model's performance among all models in solving this task. We can see that for the first 10 tasks whose training split we used to fine-tune models, the method select the model fine-tuned with corresponding dataset which is the most reasonable in semantics. And except the three Forecasting (FO) tasks, BigData22, ACL18, 
CIKM18, other tasks can achieve the best performance using the models obtained by such allocation, which also show the strangeness of these three tasks. As for the remaining tasks, the method choose the learnware model with the best performance for 4 of the 7 tasks. Specifically, the learnware model fine-tuned using the FPB dataset is selected for 5 tasks: SC, FinArg-ARC, FinArg-ACC, FOMC and MultiFin. It is worth noting that the FPB task is a sentiment analysis task under the Textual Analysis (TA) axonomy, and 4 of the 5 tasks are TA tasks, and all 5 tasks are classification tasks, which also demonstrates the method’s underlying rationality to some degree.

Here for the fist 10 tasks, we simulated a scenario that the learnware dock system can aggregate models from developers who share well-trained SLMs without revealing their private or high-quality data. In this scenario, the LDS performs significantly better than the compared 70B-scale general-purpose LLMs, verifying both the identification effectiveness of parameter vector specifications and the capabilities of fine-tuned SLMs on specifc tasks. For the remaining tasks, although there is no learnware prepared for these tasks, the experimental results show that by leveraging specialized learnwares in financial scenarios, the LDS can still outperform base SLMs, and even ourperform Qwen1.5-110B and Qwen2.5-72B, verifying the potential of identifying and reusing specialized learnwares for solving specialized tasks.

\subsection{Detailed evaluation results}
\label{sec:extra}

This section shows the remaining evaluation results of our systems. Despite the great potential, our work makes initial exploration on LLM learnware identification. The parameter vector specification currently serves as an initial attempt for identifying task-specific capabilities needed to solve user tasks. However, LLMs possess more advanced capabilities beyond this, especially reasoning in mathematics, which have not been specifically characterized in our experiments.
Actually, other post-training techniques~\citep{ouyang2022training,Fu:Peng:Ou2023}, also ultimately affect the model's parameters, allowing the parameter vector specification to capture the capabilities learned by the model in the future.

\clearpage

\begin{figure}[t]
    \centering
    \includegraphics[width=\linewidth]{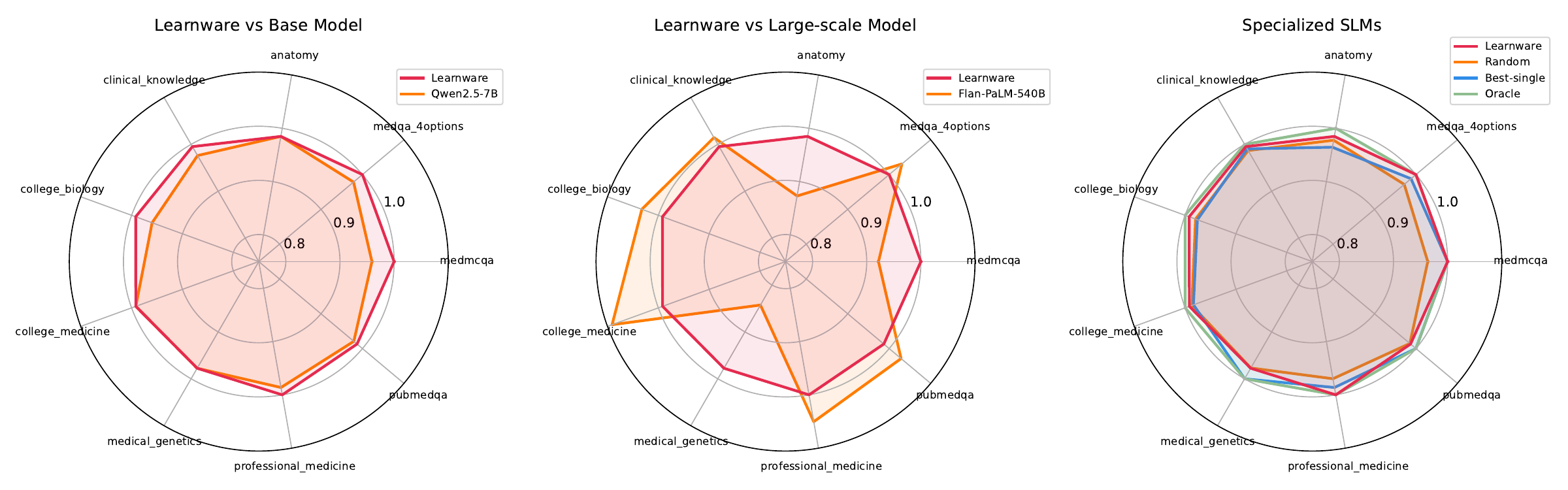}
    \caption{Performances our medical LLM evaluation benchmark. The performance metrics are also normalized relative to the ``Oracle''. Detailed performance values are shown in \Cref{tab:med}.}
    \label{fig:llm-med}
\end{figure}

\begin{figure}[t]
    \centering
    \includegraphics[width=\linewidth]{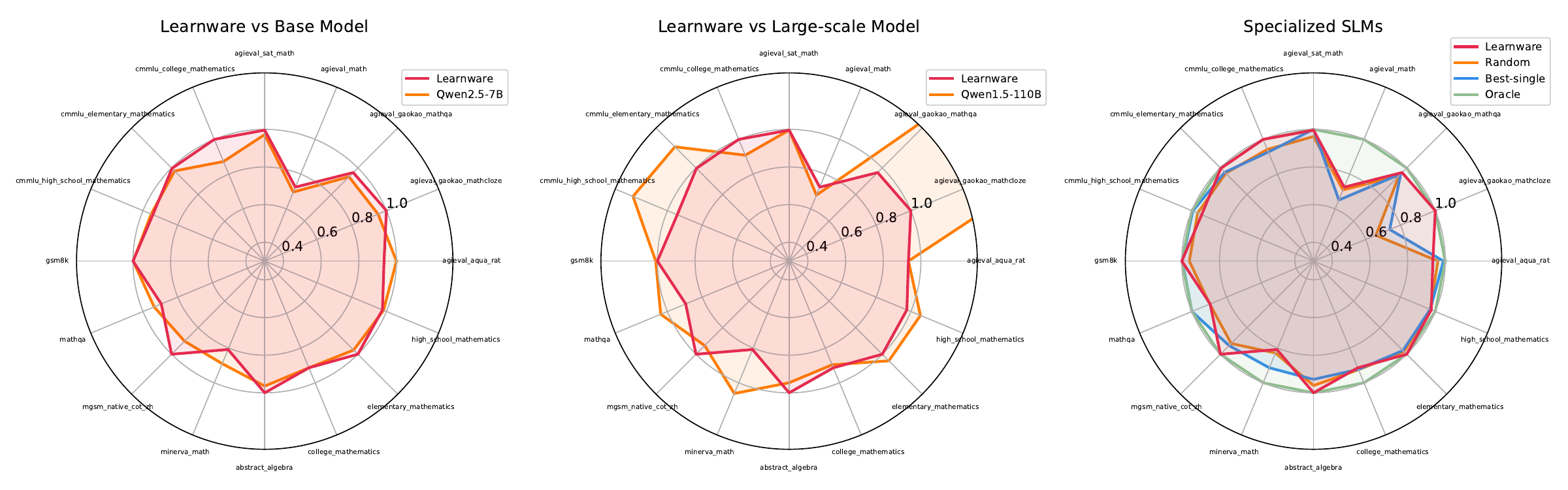}
    \caption{Performances our mathematical LLM evaluation benchmark. The performance metrics are also normalized relative to the ``Oracle''. Detailed performance values are shown in \Cref{tab:math}.}
    \label{fig:llm-math}
\end{figure}

\begin{table}[t]
\centering
\caption{Performance value of different methods or language models on our mathematical evaluation benchmark. Full score is 100. The highest score among all contenders, except oracle-style strategies \texttt{Best-Single} and \texttt{Oracle}, is emphasized in bold. ``*'' denotes the best score of all contenders in ``Specialized SLMs'' except \texttt{Oracle}.}
\label{tab:math}
\resizebox{\textwidth}{!}{
\begin{tabular}{l|cc|cc|cc}
\toprule
\multicolumn{1}{c}{User Task} & \multicolumn{1}{c}{Base Model} & \multicolumn{1}{c}{Large-scale Model} & \multicolumn{4}{c}{Specialized SLMs} \\
\cmidrule(lr){1-1} \cmidrule(lr){2-2} \cmidrule(lr){3-3} \cmidrule(lr){4-7}
Dataset                   & Qwen2.5-7B          & Qwen1.5-110B             & Random      & \textbf{Learnware}    & Best-single & Oracle         \\
\midrule
agieval\_aqua\_rat               & \textbf{41.73}      & 38.98                      & 40.09       & 38.98           & 41.34*           & 41.73 \\
agieval\_gaokao\_mathcloze       & 16.95               & \textbf{38.14}             & 11.72       & 17.80*           & 13.14          & 17.80 \\
agieval\_gaokao\_mathqa          & 49.86               & \textbf{77.78}             & 50.35       &  51.57*          & 51.00          & 53.42 \\
agieval\_math                    & 19.80               & 19.30                      & 20.15       & \textbf{20.60*}           & 18.50 & 28.40 \\
agieval\_sat\_math               & 55.91               & \textbf{57.27}                      & 55.30       &  \textbf{57.27}          & 57.50*           & 57.50 \\
cmmlu\_college\_mathematics      & 45.71               & 47.62                      & 49.36       & \textbf{52.38*}           & 48.58 & 52.38 \\
cmmlu\_elementary\_mathematics   & 65.65               & \textbf{77.83}             & 64.49       & 66.96*           & 65.00          & 67.18 \\
cmmlu\_high\_school\_mathematics & 61.59               & \textbf{77.44}             & 62.50       & 60.98           &  64.33*          & 64.63 \\
gsm8k                            & 84.08               & \textbf{84.91}             & 80.79       & 84.15*           & 83.93          & 84.15 \\
mathqa                           & 43.32               & \textbf{48.07}             & 41.51       & 41.41           &  46.28*          & 46.28 \\
mgsm\_native\_cot\_zh            & 66.40               & 68.80                      & 67.64       & \textbf{73.60*}           & 68.80 & 73.60 \\
minerva\_math                    & 40.16               & \textbf{47.90}             & 37.40       & 36.48           &  41.23*          & 45.12 \\
mmlu\_abstract\_algebra                & 54.00               & 53.00                      & 53.83       & \textbf{56.00*}           & 52.00 & 56.00 \\
mmlu\_college\_mathematics             & 53.00               & 52.00                      & \textbf{53.61*}       & 53.00           & 53.50           & 58.00 \\
mmlu\_elementary\_mathematics          & 72.75               & \textbf{78.84}             & 73.63       & 75.13*           & 73.02          & 75.13 \\
mmlu\_high\_school\_mathematics        & 55.93               & \textbf{60.00}             & 55.21       & 55.56*           &  55.19         & 56.86 \\
\midrule
Avg.                     & 51.68   & \textbf{57.99}          & 51.10     & 52.62*          & 52.08 & 54.89 \\
Avg. rank(↓) & 4.31 & \textbf{2.56} & 4.56 & 3.19* & 4.00 & - \\
\textbf{Learnware} (win/tie/loss) & 10/1/5 & 5/2/9 & 11/0/5 & - & 10/0/6 & 0/6/10 \\
Oracle (win/tie/loss) & 15/1/0 & 7/0/9 & 16/0/0 & 10/6/0 & 14/2/0 & - \\
\bottomrule
\end{tabular}
}
\end{table}

\end{document}